\documentclass[referee,pdflatex,sn-nature]{sn-jnl}

\usepackage{graphicx}%
\usepackage{subcaption} 
\usepackage{amsmath,amssymb,amsfonts}%
\usepackage{amsthm}%
\usepackage{mathrsfs}%
\usepackage[title]{appendix}%
\usepackage{xcolor}%
\usepackage{textcomp}%
\usepackage{manyfoot}%
\usepackage{booktabs}%
\usepackage{algorithm}%
\usepackage{algorithmicx}%
\usepackage{algpseudocode}%
\usepackage{listings}%
\usepackage{booktabs, makecell, multirow, tabularx}
\usepackage{threeparttable}
\usepackage{ragged2e}
\usepackage{arydshln}

\usepackage{xr}
\theoremstyle{thmstyleone}%
\theoremstyle{thmstyletwo}%

\theoremstyle{thmstylethree}%

\begin{document}

\title[Article Title]{Generating transition states of chemical reactions via distance-geometry-based flow matching}


\author[1]{\fnm{Yufei} \sur{Luo}}\email{yufei\_luo@stu.xjtu.edu.cn}

\author[1,3]{\fnm{Xiang} \sur{Gu}}\email{xianggu@xjtu.edu.cn}

\author*[1,2,3]{\fnm{Jian} \sur{Sun}}\email{jiansun@xjtu.edu.cn}

\affil[1] {\orgdiv{School of Mathematics and Statistics}, 
\orgname{Xi'an Jiaotong University}, 
\orgaddress{
\city{Xi'an}, 
\state{Shannxi Province}, 
\country{China}
}}
\affil[2]{\orgdiv{State Industry-Education Integration Center for Medical Innovations at Xi'an Jiaotong University}, 
\orgaddress{
\city{Xi'an}, 
\state{Shannxi Province}, 
\country{China}
}}
\affil[3]{\orgdiv{National Engineering Laboratory for Big Data Algorithms and Analysis Technologies}, 
\orgname{Xi'an Jiaotong University}, 
\orgaddress{
\city{Xi'an}, 
\state{Shannxi Province}, 
\country{China}
}}

\abstract{Transition states (TSs) are crucial for understanding reaction mechanisms, yet their exploration is limited by the complexity of experimental and computational approaches.
Here we propose TS-DFM, a flow matching framework that predicts TSs from reactants and products.
By operating in molecular distance geometry space, TS-DFM explicitly captures the dynamic changes of interatomic distances in chemical reactions.
A network structure named TSDVNet is designed to learn the velocity field for generating TS geometries accurately.
On the benchmark dataset Transition1X, TS-DFM outperforms the previous state-of-the-art method React-OT by 30\% in structural accuracy. 
These predicted TSs provide high-quality initial structures, accelerating the convergence of CI-NEB optimization.
Additionally, TS-DFM can identify alternative reaction paths.
In our experiments, even a more favorable TS with lower energy barrier is discovered.
Further tests on RGD1 dataset confirm its strong generalization ability on unseen molecules and reaction types, highlighting its potential for facilitating reaction exploration.
}




\maketitle
\section{Introduction}\label{sec1}
Transition states (TSs) refer to saddle points on the potential energy surface (PES) connecting reactant and product minima \cite{F29848000227}.
Predicting TS is essential for understanding chemical reactions, as the energy barrier between TS and reactant governs the reaction kinetics, while the geometric configuration of TS offers insights into the reaction mechanism \cite{eTransitionpathTheoryPathfinding2010}.
These understandings serve as the cornerstone for wide applications, including the construction of complex reaction networks \cite{https://doi.org/10.1002/wcms.1354}, design of new catalysts \cite{klucznikComputationalPredictionComplex2024, zhangExploringFrontiersCondensedphase2024}, optimization of reaction conditions \cite{durantePredictionOrganicReaction2000}, the development of novel synthetic routes \cite{kwonComputationalTransitionStateDesign2018}, etc.

The femtosecond-scale lifetime of TSs makes them difficult to characterize experimentally, and only few studies have directly captured TS structures or properties.
For instance, ultrafast electron diffraction has been used to get a direct observation of TS \cite{liuRehybridizationDynamicsPericyclic2023}.
While its high cost prevents widespread application.
Alternatively, TS exploration can be performed systematically using search algorithms combined with quantum mechanical calculations like density functional theory (DFT). 
Common TS search algorithms include growing string method \cite{10.1063/1.1691018}, nudged elastic band (NEB) method \cite{doi:10.1142/9789812839664_0016}, and Hessian-based optimization methods \cite{denzelHessianMatrixUpdate2020, hermesSellaOpenSourceAutomationFriendly2022}. 
This methodology has been applied to construct complex reaction networks \cite{zengComplexReactionProcesses2020, zhaoSimultaneouslyImprovingReaction2021}. 
However, such computations are computationally expensive and prone to convergence issues.

The emergence of machine learning has led to efficient methods for predicting TSs, overcoming the limitations of traditional approaches. 
These methods fall into two main paradigms.
The first category employs Machine Learning Interatomic Potential (MLIP) to iteratively refine initial configurations of TSs, bypassing the need for expensive DFT calculations.
NeuralNEB \cite{Schreiner_2022} trains an MLIP for the NEB algorithm to approximate the minimum energy path.
MLHessian-TSOpt \cite{yuanAnalyticalInitioHessian2024a} trains a fully differentiable MLIP NewtonNet, which is then used to predict analytical Hessians for saddle point optimization to find TS.
Another mainstream approaches utilize machine learning to predict TS directly.
Early works generate TSs in one-step manner.
For instance, TS-Gen \cite{D0CP04670A} employs a graph neural network to predict the distance matrix, followed by least-squares optimization to reconstruct the TS coordinates.
TSNet \cite{D1SC01206A} proposes an end-to-end network structure to predict TS coordinates directly. 
However, their experiments are limited to specific reaction types. 
Beyond these, TS-GAN \cite{10.1063/5.0055094} adopts a generative adversarial network (GAN) to map the reactant–product space and generate TS distance matrix.
Choi \cite{choiPredictionTransitionState2023a} proposes a PSI-based model for predicting TSs of general organic reactions, in which the interatomic distances of TS are derived from atomic pair features reflecting reactant, product, and linear interpolation of their Cartesian coordinates.
Recently, multi-step generation methods are also utilized for generating TS, achieving state-of-the-art results.
OA-ReactDiff \cite{duanAccurateTransitionState2023a} develops an object-aware SE(3)-equivariant diffusion model that satisfies all physical symmetries and constraints for generating sets of reactant, TS, and product structures.
However, an additional confidence model is required to select the best TS from randomly generated samples.
To fix this, React-OT \cite{duanOptimalTransportGenerating2025a} utilizes an optimal transport approach for generating unique TS from reactant and product.

While recent generative approaches like OA-ReactDiff and React-OT achieve notable success in generating TS, their operation in Cartesian coordinate space obscures the essence of chemical reactions, i.e., the dynamic changes in interatomic distances.
This might lead to inaccurate predictions and hinder the generalization ability on unseen reactions.
To directly target the core of chemical reactions, we propose a novel framework for \textbf{T}ransition \textbf{S}tate prediction via optimal transport conditional \textbf{F}low \textbf{M}atching in molecular \textbf{D}istance geometry space, dubbed TS-DFM.
In TS-DFM, the task of TS prediction is formulated as fitting a linear velocity field derived from the optimal transport map and conditioned on the reactant and product, which directs the initial guessed TS toward the corresponding true TS.
Using this learned velocity field, TSs are predicted by numerically solving an ordinary differential equation starting from the initial guess, followed by nonlinear optimization to recover the Cartesian coordinates of the TS.
To generate more chemically plausible starting points and improve prediction accuracy, the initial guessed TSs are constructed by distance-geometry interpolation inspired by IDPP \cite{10.1063/1.4878664}, avoiding unphysical bond distortions inherent in Cartesian interpolation as used in the React-OT and PSI-based models.
Additionally, we design a two-branch network structure named TSDVNet for fitting the linear velocity field of the optimal transport map.
The architecture employs an auxiliary branch to process the conditional input, whose produced representations are fused with the main branch to improve the prediction of the velocity field.

Our proposed TS-DFM achieves 30\% higher average structural accuracy than SOTA method React-OT on Transition1x dataset \cite{schreinerTransition1xDatasetBuilding2022a}.
These structures could also serve as better initial configurations for subsequent Climbing Image NEB (CI-NEB) optimization, converging three times faster than IDPP initialization method and at least 10\% faster than compared TS prediction approaches.
TS-DFM also enables the discovery of various possible reaction paths through normal mode sampling on reactant and product structures.
In our experiments, we even discover a more favorable TS with lower energy barrier for one specific reaction.
Furthermore, TS-DFM exhibits better generalization ability than React-OT, outperforming it by at least 16\% on average RMSD on unseen reaction types and molecular structures.

\section{Results}\label{sec2}
\subsection{Overview of TS-DFM} 
Current state-of-the-art TS prediction methods have adopted multi-step generative paradigms (e.g., diffusion or flow matching) in Cartesian coordinate space.
However, operating in this space obscures the intrinsic bonding evolution during chemical reactions.
Although predicting TSs through pairwise distances, as done in some earlier works, provides a more intrinsic representation, these methods face limitations including suboptimal network design and one-step prediction scheme that underperforms these modern generative paradigms.
Therefore, we propose a novel framework named TS-DFM, which utilizes optimal transport conditional flow matching in molecular distance geometry space to predict TSs.
This solution explicitly models the bonding evolution in the molecular distance geometry space, representing via pairwise distance matrices, to better capture the essentials of chemical reactions.

\begin{figure}[ht]
    \centering
    \includegraphics[width=\textwidth]{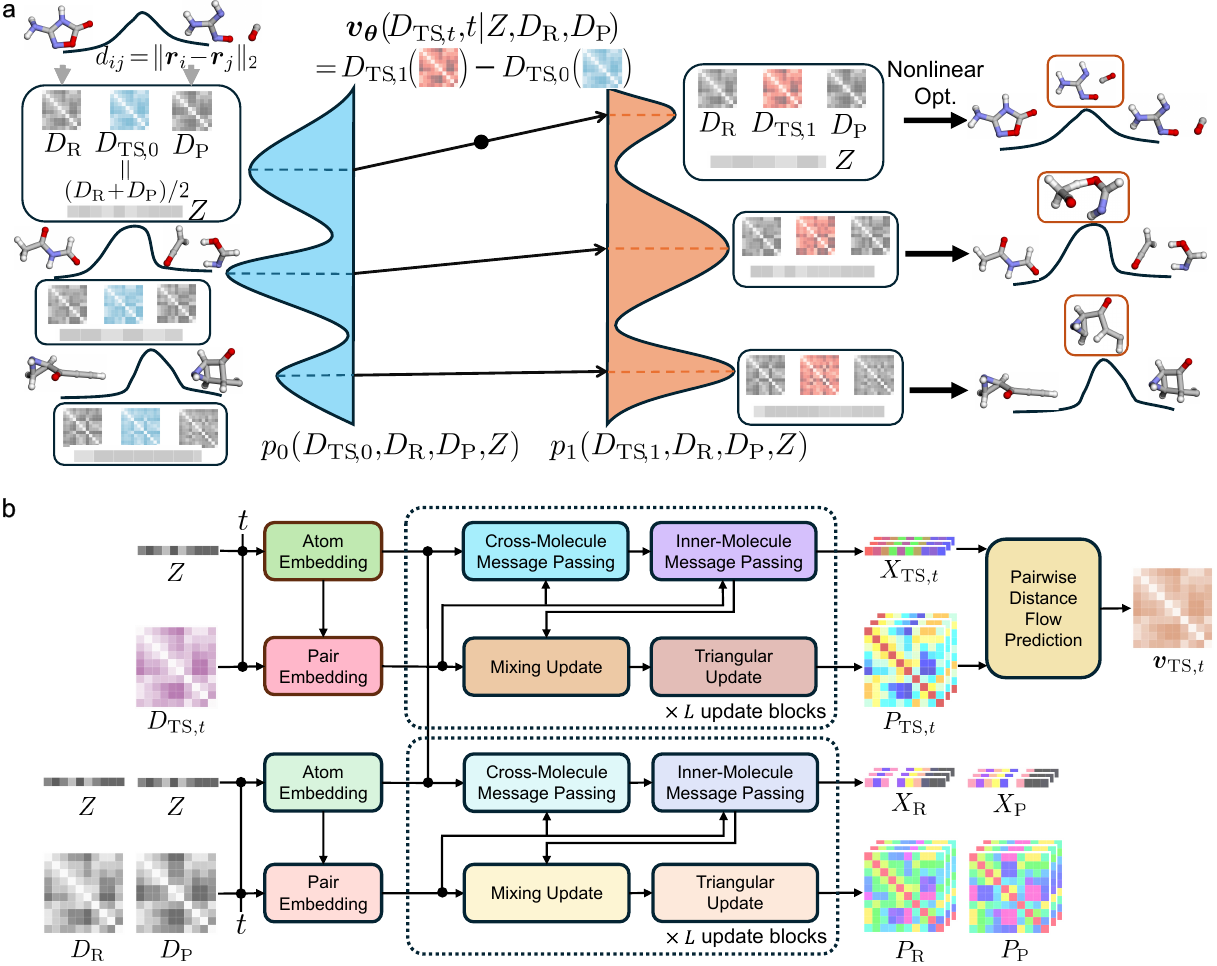}
    
    \caption{\textbf{Overview of TS-DFM and TSDVNet}.
    \textbf{a}, Schematic representations of TS-DFM.
    Let $p_0$ and $p_1$ denote the data distributions over the initial guessed and true TSs of chemical reactions, respectively.
    Initial guessed and true TSs from the same reaction are coupled. 
    We learn a linear velocity field in the distance geometry space to evolve an initial guessed TS toward the corresponding true TS.
    \textbf{b}, The network structure TSDVNet for learning the velocity field.
    The network consists of two branches with identical architecture but independent parameters. 
    The upper branch predicts the velocity field, while the lower branch encodes the representations of reactant and product.
    Features from both branches are integrated by feature fusion operations to produce enriched representations.
    }
    \label{fig:overview&structure}
\end{figure}

The idea is to first map the reactant and product structure to the pairwise distance matrices $D_{\text{R}}$ and $D_{\text{P}}$.
Then we design a conditional flow matching framework to estimate the distance matrix of TS conditioned on $D_{\text{R}}$, $D_{\text{P}}$, and atom types $Z$. 
Instead of starting from the generally utilized Gaussian noise in flow matching model, TS-DFM takes $(D_{\text{R}} + D_{\text{P}})/2$ as the starting point to avoid unreasonable bond lengths in Cartesian linear interpolation and improve both the efficiency and accuracy.
After predicting the distance matrix of TS, its atomic coordinates are reconstructed by nonlinear optimization.
This framework is illustrated in Fig. \ref{fig:overview&structure}a, in which we demonstrate the learned velocity field $v_{\boldsymbol{\theta}}$ transports the source distribution $p_0$ denoting the initial guess to the target distribution $p_1$ representing the TS.

Under this framework, the major contribution is to design the optimal transport based conditional flow matching model to generate the distance matrix of TS, conditioned on the reactant and product.
To implement distance geometry evolution, we designed a network named TSDVNet to model the velocity field that drives the initial distance matrix toward that of TS.
As illustrated in Fig. \ref{fig:overview&structure}b, the network takes $D_{\text{R}}$, $D_{\text{P}}$ and $Z$ as conditional inputs. 
Their representations, generated via an additional branch, are fused into the main branch to facilitate the prediction of the velocity field.
We here just briefly describe the model and network design of TS-DFM, and the details are presented in ``Details about TS-DFM'' and ``Network Structure of TSDVNet'' respectively.
In the following sections, we will evaluate TS-DFM by demonstrating its superior prediction accuracy and generalization ability, and exploring its capabilities in accelerating CI-NEB optimization and uncovering new reaction paths.

\subsection{Generating High-quality TS from TS-DFM}
We first trained TS-DFM on Transition1x dataset.
The data is generated by performing CI-NEB calculations with DFT at the $\omega$B97x/6-31G(d) level on 10k organic reactions, resulting in 9.6 million calculated energies and forces for molecular configurations on and around the reaction pathways.
The reactant, product and TS are extracted from the converged path of CI-NEB calculations.
In our experiments, we use the default data split in the officially released database.

To validate the effectiveness of TS-DFM, we compare it with the following four representative baseline approaches.
NeuralNEB \cite{Schreiner_2022} utilizes a MLIP as a surrogate PES for NEB calculation.
In NEB calculations, we adopt the settings in \cite{Schreiner_2022}, utilizing IDPP for initialization and the climbing-image method for iterative optimization.
The details of MLIP training and NEB configurations are elaborated in ``Implementation Details''.
PSI-based model \cite{choiPredictionTransitionState2023a} has the best performance among one-step generation methods. 
OA-ReactDiff \cite{duanAccurateTransitionState2023a} is a representative approach that generates TSs randomly.
For comparison, we follow \cite{duanAccurateTransitionState2023a, duanOptimalTransportGenerating2025a} to sample 20 structures for each test data and keep the one with the lowest RMSD.
React-OT \cite{duanOptimalTransportGenerating2025a} achieves SOTA results among the existing TS prediction methods.

We compare the performance of different methods on evaluation metrics that fall into two categories: structural error and potential energy properties.
The structural error, measured by RMSD and DMAE, assesses the geometric deviation of the predicted TS.
The second category evaluates the potential energy properties derived from DFT calculations on the predicted TS. 
This includes the absolute error in TS energy $|\Delta E_{\text{TS}}|$, the maximum and root-mean-square of atomic forces, i.e., $f_{\max}$ and $F_{\text{rms}}$. 
As a saddle point, a valid TS must exhibit near-zero forces.
Additionally, harmonic analysis is performed based on the Hessian matrix, and the predicted structure with only one imaginary frequency is identified as a saddle point.
The detailed description and mathematical definitions of these metrics are in ``Evaluation Metrics''.

In comparison with baseline methods, TS-DFM achieves the best overall performance. 
It demonstrates superior structural accuracy, with a mean RMSD of 0.1828 \AA ~and a mean DMAE of 0.0607 \AA, outperforming the best baseline React-OT by approximately 30\% on both metrics.
Notably, TS-DFM achieves this with only half model parameters (5.2M versus 10.6M).
TS-DFM also has the lowest average $|\Delta E_{\text{TS}}|$ of 0.1614 eV, and 65.5\% of its predicted TSs are located at saddle points, both of which surpass the second best method NeuralNEB (0.2970 eV and 63.41\% respectively).
By iteratively refining image configurations toward saddle points in NEB calculations, NeuralNEB achieves the best performance in force-related metrics ($f_{\max}$: 1.1190 eV/\AA; $F_{\text{rms}}$: 0.5588 eV/\AA).
TS-DFM performs most closely to it ($f_{\max}$: 1.6482 eV/\AA; $F_{\text{rms}}$: 0.8227 eV/\AA), significantly outperforming the other three methods. 
In terms of efficiency, TS-DFM generates TS in only 1.2 seconds on average (three times slower than React-OT but 50 times faster than NeuralNEB), achieving a practical trade-off between speed and accuracy. 
We further conduct two ablation studies to investigate the source of TS-DFM's superior performance (see Supplementary Tab. \ref{supp_tab:ablation}).
The results demonstrate the effectiveness of TSDVNet, linear interpolation in pairwise distance space, and flow matching objective all contribute to the superior performance of TS-DFM.

\begin{figure}[!ht]
    \centering
    \includegraphics[width=\textwidth]{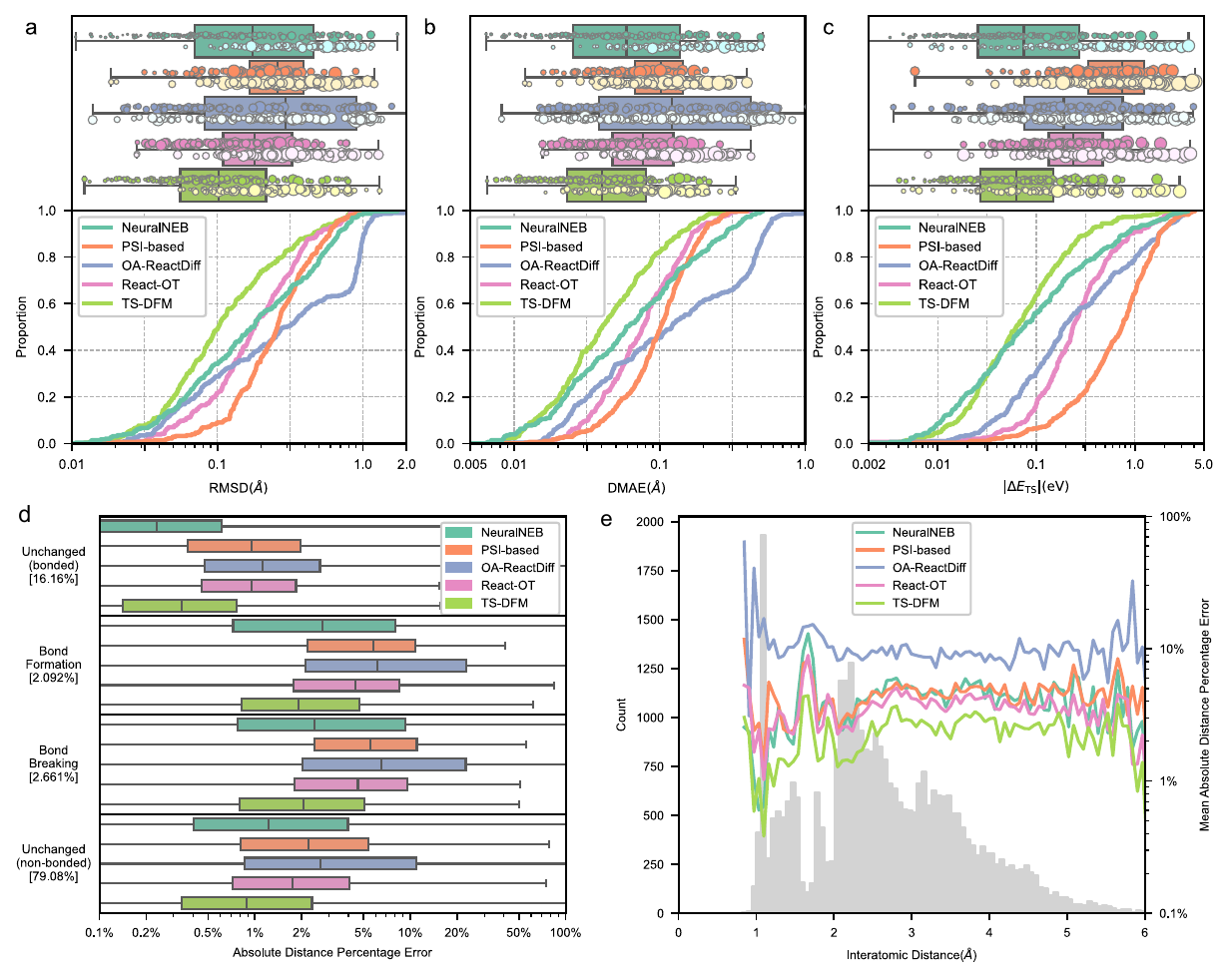}
    \caption{\textbf{Error analysis of different methods}.
    \textbf{a}, \textbf{b} and \textbf{c}, The boxplots (upper) and cumulative probabilities (lower) for RMSD, DMAE and $|\Delta E_{\text{TS}}|$ of the predicted TSs on test dataset respectively.
    The prediction error for each test data is also plotted as a point on the boxplot, where the point’s size is proportional to its $f_{\max}$ value, its color (dark or light) denotes whether the predicted TS lies at a saddle point or not.
    \textbf{d}, Boxplots for the absolute percentage errors in pairwise distances of predicted TSs, categorized by the type of bond change during the chemical reaction.
    Percentages in brackets indicate the proportion of each bond change type within the entire test set.
    \textbf{e}, Average absolute percentage errors and numbers of atom pairs w.r.t. pairwise distances.
    The gray histogram shows the distribution of interatomic distances.
    The colored line represents the mean absolute percentage error computed over different distance intervals.
    }
    \label{fig:res_ts1x_orig}
\end{figure}

Figure \ref{fig:res_ts1x_orig}a-c illustrates the distribution of different evaluation metrics in the test set.
TS-DFM achieves the best cumulative probabilities of RMSD and DMAE among the compared methods.
For $|\Delta E_{\text{TS}}|$, NeuralNEB performs the best at values below approximately the 40th percentile of the error distribution, while TS-DFM surpasses other methods at values above this threshold.
To determine whether TS-DFM better captures the changes in interatomic distances, we further compare the distributions of absolute percentage errors in the interatomic distances of predicted TSs across different methods, as shown in Fig. \ref{fig:res_ts1x_orig}e and f.
In Fig. \ref{fig:res_ts1x_orig}e, we apply the method proposed in \cite{https://doi.org/10.1002/bkcs.10334} to infer chemical bonds in reactants and products. 
Atom pairs are classified into four categories based on their bonding evolution during the reactions.
Among five approaches, NeuralNEB performs best in predicting distances of unchanged chemical bonds, while TS-DFM achieves the highest accuracy for the other three bond-change types. 
Notably, for bond breaking and formation—the key processes in chemical reactions, TS-DFM slightly outperforms NeuralNEB and shows at least 2\% improvement over the other three methods in the median values.
We also analyze the average distance percentage error as a function of interatomic distance, with results presented in Fig. \ref{fig:res_ts1x_orig}f. 
These results demonstrate that TS-DFM performs the best across almost all interatomic distances. 
The weakest performance occurs at approximately 1.6~\AA, a distance that corresponds to the typical range of chemical bond breaking and formation. 
The significant errors in this region can likely be attributed to data scarcity, as this is commonly observed in other data-limited regions.
In conclusion, TS-DFM more effectively captures the changes in interatomic distances during chemical reactions compared to previous methods, leading to superior accuracy in predicting TSs.

\subsection{TS-DFM Provides Better Initialization for TS Search Algorithms}
As illustrated in the previous section, the predicted TSs from TS-DFM still have gaps to the saddle points.
Although not perfectly accurate, the predicted structures offer useful initial configurations that enhance convergence in TS search algorithms.
In this section, we take NEB calculation as an example to evaluate their suitability as starting configurations.
Due to the high computational cost of DFT, MLIP is used as a surrogate PES.
The traditional IDPP interpolation is set as a baseline initialization method.
For PSI-based model, OA-ReactDiff, React-OT, and our proposed TS-DFM, the predicted TS from each method serves as an additional anchor point in NEB images between reactant and product.
Intermediate images are then placed along reactant-TS and TS-product segments, followed by IDPP initialization.

To comprehensively evaluate the performance of different approaches, Tab.~\ref{tab:combined_results} presents a comparative analysis of both directly predicted and CI-NEB optimized TSs.
The upper part summarizes the results of direct prediction methods mentioned in the previous section, while the lower part demonstrates the performance after CI-NEB optimization.
This integrated comparison allows a clear assessment of how initialization methods influence the quality of optimized TSs.
In comparison with the results from upper and lower parts of Tab.\ref{tab:combined_results}, the optimized TSs from PSI-based model, OA-ReactDiff and React-OT demonstrate significant refinements upon their directly predicted structures.
However, OA-ReactDiff suffers from severe convergence problem in subsequent NEB calculations, where only 59.23\% of test data successfully converged in NEB calculations and exhibits the worst average performance.
The rest of the approaches all surpass IDPP, and our proposed TS-DFM achieves the best results across all metrics.
For TS-DFM, the structure refinements are mainly represented in the decrease of $f_{\max}$ and $F_{\text{rms}}$, with only a slight change in RMSD, DMAE and $|\Delta E_{\text{TS}}|$.
Similar to Fig. \ref{fig:res_ts1x_orig}, the distributions and cumulative probabilities of different metrics after NEB optimization are illustrated in Fig. \ref{suppl_fig:res_ts1x_neb}.
TS-DFM achieves the best cumulative probabilities on three metrics among the compared initialization methods.
Additionally, TS-DFM has the best convergence time on average, outperforming other initialization method by at least 10\%.
Therefore, the predicted TS of TS-DFM needs less iterations in NEB calculation than the compared methods.
The longer inference time of TS-DFM is compensated by its faster NEB convergence, making it the most efficient method overall.
In practice, this could reduce much computational burden when TS with quantum chemistry accuracy is required.

\begin{table*}[ht]
  \centering
  \caption{Statistics of evaluation metrics for the predicted TSs from different approaches and CI-NEB optimized TSs from different initialization methods. The best results are shown in boldface, and the second best are underlined.}
   \resizebox{\textwidth}{!}{
   \renewcommand{\arraystretch}{1.2}
    \begin{tabular}{l|l|cccccccccccccc}
    \hline
    \multirow{2}[1]{*}{\makecell{Method \\ Type}} & \multirow{2}[1]{*}{\makecell{Approaches/ \\[-2pt] Init. method}} & \multirow{2}[1]{*}{Params} & \multirow{2}[1]{*}{\makecell{Time\\(s)}} & \multirow{2}[1]{*}{\makecell{\% of NEB \\[-2pt] convergence}} & \multirow{2}[1]{*}{\makecell{\% of \\[-2pt] saddle pts}} &  \multicolumn{2}{c}{RMSD (\AA)} & \multicolumn{2}{c}{DMAE (\AA)} & \multicolumn{2}{c}{$|\Delta E_{\text{TS}}|$ (eV)} & \multicolumn{2}{c}{$f_{\max}$ (eV/\AA)} & \multicolumn{2}{c}{$F_\text{rms}$ (eV/\AA)} \\
           & & & & & & mean & median & mean  & median & mean  & median & mean  & median & mean  & median \\
    \hline
    \hline
    \multirow{5}{*}{\makecell{Direct \\ Prediction}} 
    & PSI-based & 11.9M & 0.1 & - & 28.70\% & 0.3035 & 0.2594 & 0.1126 & 0.1019 & 0.9191 & 0.7383 & 4.4587 & 3.8852 & 2.1864 & 1.9574 \\
    & OA-ReactDiff & 10.6M & 5.1 & - & 40.40\% & 0.4845 & 0.3043 & 0.2259 & 0.1217 & 0.5979 & 0.1912 & 2.6850 & 2.3206 & 1.3356 & 1.1764 \\
    & React-OT & 10.6M & 0.4 & - & \underline{55.60\%} & \underline{0.2436} & \underline{0.1803} & \underline{0.0934} & \underline{0.0766} & \underline{0.4272} & \underline{0.2359} & \underline{3.4365} & \underline{2.9620} & \underline{1.7010} & \underline{1.5123} \\
    & TS-DFM & 5.2M & 1.2 & - & \textbf{65.50\%} & \textbf{0.1828} & \textbf{0.1000} & \textbf{0.0607} & \textbf{0.0402} & \textbf{0.1614} & \textbf{0.0617} & \textbf{1.6482} & \textbf{1.2259} & \textbf{0.8227} & \textbf{0.6100} \\
    \hline
    \hline
    \multirow{5}{*}{\makecell{CI-NEB \\ Optimized}} 
    & IDPP(NeuralNEB) & 5.0M & 70.0 & 88.50\% & 63.41\% & 0.2836 & 0.1756 & 0.1066 & 0.0592 & 0.2970 & 0.0748 & 1.1190 & 0.7018 & 0.5588 & 0.3598 \\
    & PSI-based & 5.0M+11.9M & 33.9 & 86.41\% & 67.25\% & 0.2332 & 0.1371 & 0.0835 & 0.0512 & 0.2076 & 0.0645 & 1.0206 & 0.6250 & 0.5177 & 0.3293 \\
    & OA-ReactDiff & 5.0M+10.6M & 28.4 & 59.23\% & 60.98\% & 0.2944 & 0.1588 & 0.1340 & 0.0562 & 0.6885 & 0.0717 & 1.4550 & 0.7592 & 0.7353 & 0.4019 \\
    & React-OT & 5.0M+10.6M & 22.9 & \underline{90.59\%} & \underline{70.38\%} & \underline{0.2049} & \underline{0.1091} & \underline{0.0743} & \underline{0.0423} & \underline{0.1712} & \underline{0.0499} & \underline{0.9803} & \underline{0.6153} & \underline{0.4851} & \underline{0.3080} \\
    & TS-DFM & 5.0M+5.2M & 19.7 & \textbf{91.29\%} & \textbf{74.22\%} & \textbf{0.1832} & \textbf{0.0862} & \textbf{0.0693} & \textbf{0.0352} & \textbf{0.1413} & \textbf{0.0444} & \textbf{0.9166} & \textbf{0.6014} & \textbf{0.4607} & \textbf{0.3024} \\
    \hline
    \end{tabular}}
    \label{tab:combined_results}
\end{table*}

\subsection{Discovering Alternative Reaction Pathways}
Chemical reactions with multiple TSs are frequently observed in polyatomic systems \cite{gimarcShapesSimplePolyatomic1970, tishchenkoUnifiedPerspectiveHydrogen2008}.
Targeting the lowest barrier TS is a burdensome task, which is often accomplished by exploring multiple TSs in practice.
In this section, we focus on exploring the ability of TS-DFM to generate multiple TSs, which is useful in determining the most favorable reaction pathway.
Considering that TS-DFM generates the TSs of chemical reactions in a deterministic manner, one possible way of generating diverse TSs is to randomly perturb the reactants and products as performed in \cite{choiPredictionTransitionState2023a}.
To ensure these perturbed structures are physically reasonable, we utilize normal mode sampling (NMS) \cite{smithANI1DataSet2017a} (See ``Implementation Details'' in Methods Section) for reactants and products.

\begin{figure}[!ht]
    \centering
    \includegraphics[width=\linewidth]{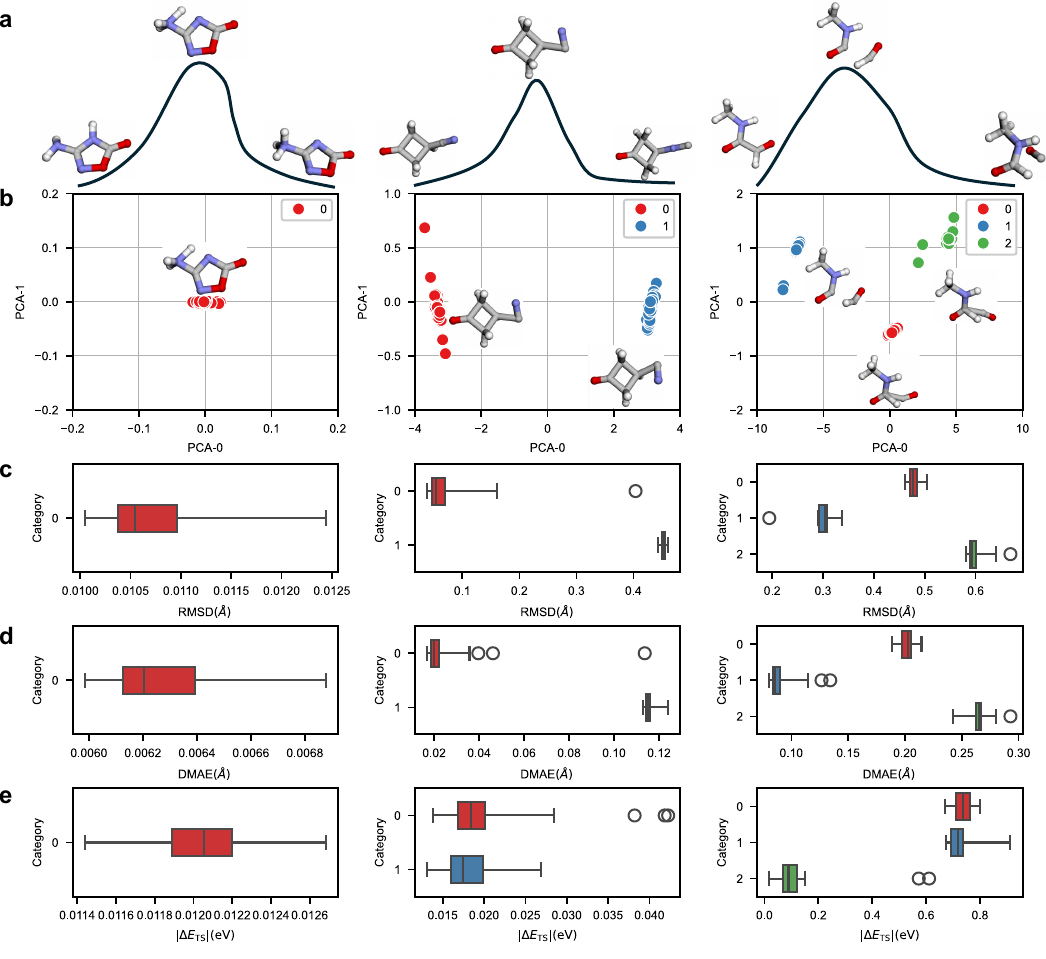}
    \caption{\textbf{Analysis of samples generated in TS exploration}. 
    \textbf{a}, Illustration of three chemical reactions in the test set.
    The H, C, N and O atoms are colored as white, gray, blue and red respectively.
    \textbf{b}, Principal component analysis (PCA) on the Coulomb matrices of 100 generated TSs.
    Each structure is colored by its K-means cluster in the PCA space.
    The structure with the lowest potential energy within each cluster is also plotted nearby.
    \textbf{c}, \textbf{d} and \textbf{e}, The boxplots of RMSD, DMAE and $|\Delta E_{\text{TS}}|$ between the generated samples and referenced TS for each cluster, respectively.}
    \label{fig:rand_gen}
\end{figure}

We investigate three different reactions from the test set of Transition1x as examples, as illustrated in Fig. \ref{fig:rand_gen}a.
Molecular structures are visualized via py3DMol \cite{10.1093/bioinformatics/btu829}.
For each of them, the reactants and products are randomly perturbed by NMS to generate different predicted structures.
Then, these structures are taken as the initial structures for CI-NEB optimization, with MLIP being used as surrogate PES.
For each reaction, we collect 100 different TSs that converged in CI-NEB calculation.
To visualize the differences of these collected TSs, they are further represented by the Coulomb matrix \cite{PhysRevLett.108.058301} followed by principal component analysis (PCA).
Figure \ref{fig:rand_gen}b shows the 2D plots of these TSs using the first two dimensions of PCA, where K-means algorithm are utilized to cluster similar structures in PCA space.
The result demonstrates that three reactions have different numbers of distinct clusters, each of which corresponds to a specific TS.
For simplicity, we denote the TS corresponds to cluster $i$ as $\text{TS-}i$.
Then, we plot the RMSD, DMAE and $|\Delta E_{\text{TS}}|$ w.r.t. the category of cluster, whose results are shown in Fig. \ref{fig:rand_gen}c, d and e respectively.
For the leftmost reaction in Fig. \ref{fig:rand_gen}, only one TS is found, whose structure is close to the reference TS.
Two TSs with close energy barriers are found for the reaction in the middle of Fig. \ref{fig:rand_gen}.
In the rightmost case of Fig. \ref{fig:rand_gen}, $\text{TS-}1$ is closest to the reference.
Interestingly, $\text{TS-}0$ and $\text{TS-}2$ correspond to new TSs with different reaction mechanisms from the reference.
Moreover, the statistics of $|\Delta E_{\text{TS}}|$ for $\text{TS-}2$ outperform $\text{TS-}1$, indicating that a TS with lower energy barrier might be found.

To confirm the validity of these possible TSs, we further perform Intrinsic Reaction Coordinate (IRC) calculations for these TSs, using DFT at $\omega$B97x/6-31G(d) level of theory.
Due to the inaccuracy of MLIP, these structures are first refined via saddle point optimization before executing IRC calculations.
The details of these calculations are in the ``Implementation Details''.
It is confirmed that all TSs shown in Fig. \ref{fig:rand_gen}b could obtain the intended reactant and product molecules in IRC calculations (see Supplementary Fig. \ref{supp_fig:rand_gen_irc}).
Specifically, for the cases in the middle of Fig. \ref{fig:rand_gen}, the refined structures of $\text{TS-}1$ has potential energy close to that of $\text{TS-}0$.
Compared to the reference structure, the reactant geometry derived from IRC calculation of $\text{TS-}1$ is a distinct conformer of the same molecule.
In the rightmost case of Fig. \ref{fig:rand_gen}, the optimized structure of $\text{TS-}1$ is almost identical to the reference TS.
However, the optimized structure of $\text{TS-}0$ is close to that of $\text{TS-}2$.
If MLIP is used in IRC calculations instead, $\text{TS-}0$ is located at the saddle point and the intended reactant and product structures are obtained, indicating that $\text{TS-}0$ is an artifact induced by the inaccuracy of MLIP.
Then, it is confirmed that the reference reactant and product molecules could be reached in IRC calculations using the optimized structure of $\text{TS-}1$ and $\text{TS-}2$.
In addition, the potential energy of the optimized $\text{TS-}2$ is 0.47eV lower than that of $\text{TS-}1$, demonstrating that a better reaction path is found.
Therefore, by additionally utilizing NMS sampling and MLIP-based CI-NEB calculation, TS-DFM exhibits the capability of exploring alternative TSs, which helps to provide comprehensive understanding of chemical reactions.

\subsection{Superior Generalization Ability to Unseen Reactions}
To explore the capability of TS-DFM, we conduct additional experiments using the RGD1 dataset \cite{zhaoComprehensiveExplorationGraphically2023a}, which contains 176k organic reactions at the B3LYP-D3/TZVP level and offers a wider variety of reaction types and molecular structures than Transition1x.
The data is partitioned into five subsets: Train-id (20k), Valid-id (1k), Test-id (1k), Test-ood-type (1k), and Test-ood-size (1k).
The splitting procedure is illustrated in Supplementary Fig. \ref{supp_fig:rgd1_split}.
In comparison with Test-id subset, Test-ood-size contains reaction data with larger molecular size, while Test-ood-type contains reaction data with different reaction type.
Therefore, this dataset can be used not only to evaluate model performance on complex chemical reactions, but also to assess generalization to unseen molecular structures and reaction types.

Table \ref{tab:res_rgd1} presents a comparative analysis between TS-DFM and React-OT using our partitioned subset of RGD1.
The results demonstrate statistically significant superiority of TS-DFM across three test subsets.
Specifically, TS-DFM achieves lower mean RMSD and DMAE values, with at least 16.0\% reductions.
This superiority is quantified by the positive $\Delta \epsilon$ values, indicating that TS-DFM obtains lower errors than React-OT in a majority of test cases (68.5\% to 86.2\%).
The consistency of this advantage is further validated by extremely low $p$-values ($p\le 9.25\times 10^{-43}$) in the Wilcoxon two-sided signed-rank tests, firmly rejecting the null hypothesis of comparable performance.
We also conduct bond analysis for the predicted TSs similar to that in Fig. \ref{fig:res_ts1x_orig}d and e, which are presented in the Supplementary Fig. \ref{supp_fig:rgd1_bond_analysis1} and \ref{supp_fig:rgd1_bond_analysis2}.
The results demonstrate that TS-DFM surpasses React-OT in different bond change types and in almost all interatomic distances. 
\begin{table*}[!ht]
  \centering
  \caption{Statistics of TS-DFM versus React-OT on RGD1 dataset.}
  \resizebox{\textwidth}{!}{
  \renewcommand{\arraystretch}{1.2}
  \begin{threeparttable}
    \begin{tabular}{lcccccccccc}
    \hline
     \multirow{2}[2]{*}{Data split} &  \multirow{2}[2]{*}{Metrics}   & \multicolumn{2}{c}{React-OT} & \multicolumn{2}{c}{TS-DFM} & \multicolumn{3}{c}{Statistics of $\Delta \epsilon$ \tnote{1}} & \multicolumn{2}{c}{Wilcoxon signed-rank test } \\
          &       &  mean  & median & mean  & median & \% of $\Delta\epsilon>0$ & $\Delta\epsilon_\text{mean}$ & $\Delta\epsilon_\text{std}$ & \makecell{$p$-value \\ (two-sided) }& \makecell{Rank biserial \\ correlation} \\
    \hline
    \multirow{2}[2]{*}{Test-id} & RMSD(\AA)  & 0.5990 & 0.5471 & \textbf{0.4395} & \textbf{0.3559} & 74.7\% & 0.1595 & 0.1528 & $1.91\times 10^{-64}$ & 0.6188 \\
          & DMAE(\AA) & 0.2191 & 0.1997 & \textbf{0.1435} & \textbf{0.1146} & 82.8\% & 0.0756 & 0.0726 & $1.23\times 10^{-101}$ & 0.7814 \\
    \hline
    \multirow{2}[2]{*}{Test-ood-size} & RMSD(\AA) & 0.6610 & 0.5983 & \textbf{0.4936} & \textbf{0.3908}  & 75.4\% & 0.1676 & 0.1583 & $2.49\times 10^{-64}$ & 0.6182 \\
          & DMAE(\AA) & 0.2480 & 0.2326 & \textbf{0.1605} & \textbf{0.1329} & 86.2\% & 0.0875 & 0.0856 & $8.90\times 10^{-117}$ & 0.8386 \\
    \hline
    \multirow{2}[2]{*}{Test-ood-type} & RMSD(\AA) & 0.7989 & 0.7492 & \textbf{0.6703} & \textbf{0.6194} & 68.5\% & 0.1286 & 0.1336 & $9.25\times 10^{-43}$ & 0.5004 \\
          & DMAE(\AA) & 0.3490 & 0.3354 & \textbf{0.2491} & \textbf{0.2203}  & 79.7\% & 0.0999 & 0.0950 & $1.70\times 10^{-91}$ & 0.7406 \\
    \hline
    \end{tabular}%
    \begin{tablenotes}
      \item[1] $\Delta \epsilon=\epsilon_{\text{React-OT}}-\epsilon_{\text{TS-DFM}}$, where $\epsilon$ denotes RMSD or DMAE of different approaches.
    \end{tablenotes}
  \end{threeparttable}}
  \label{tab:res_rgd1}%
\end{table*}%

\begin{figure}[!ht]
    \centering
    \includegraphics[width=\linewidth]{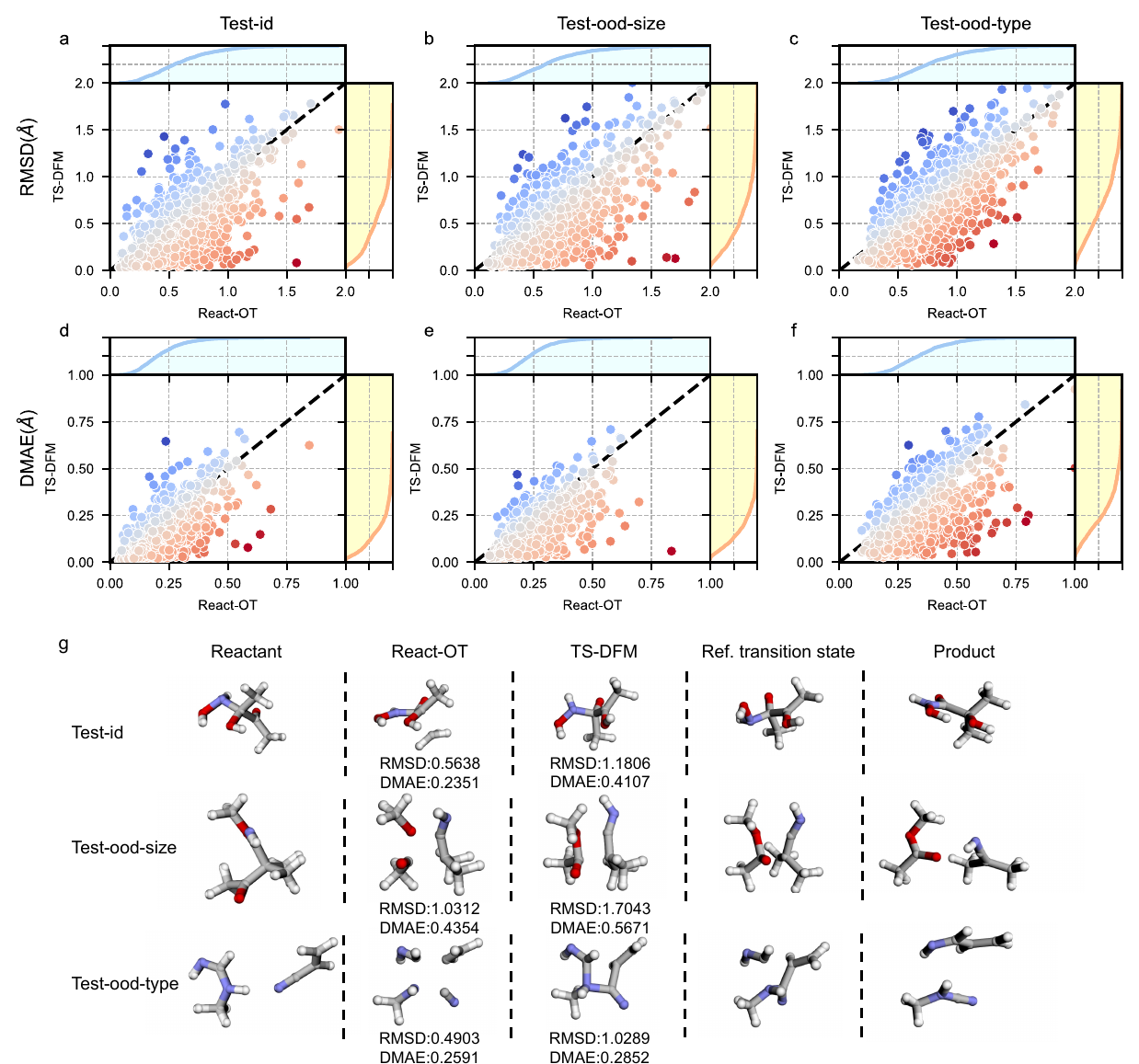}
    \caption{\textbf{Performance comparison and case studies of TS-DFM versus React-OT on different test subsets}.
    \textbf{a}, \textbf{b} and \textbf{c}, The RMSD of TS-DFM versus React-OT for each data on Test-id, Test-ood-type and Test-ood-size subsets respectively.
    \textbf{d}, \textbf{e} and \textbf{f}, The corresponding comparisons for DMAE.
    The x and y axis correspond to the RMSD or DMAE of React-OT and TS-DFM.
    On top and right of each subfigure are the respective cumulative probability distribution of React-OT and TS-DFM.
    A blue-white-red color gradient is used to represent the difference in RMSD or DMAE between the two methods, where darker blue indicates that React-OT performs better, darker red indicates that TS-DFM performs better, and white suggests that the metrics of the two methods are close.
    \textbf{g}, Examples of TS-DFM underperform React-OT on different test subsets.
    The gray, blue, red, and white spheres represent C, N, O, and H respectively.
    Both methods exhibit limitations such as inaccurate torsion angles, misoriented substructures, erroneous bond predictions, and even incorrect reaction mechanism.
    Nevertheless, even in some of the worst cases, TS-DFM could still demonstrate partially (e.g., correct bond changes) reaction mechanisms.
    }
    \label{fig:rgd1_compare}
\end{figure}

A comparative demonstration for React-OT versus TS-DFM on different subsets is represented in Fig \ref{fig:rgd1_compare}a-f.
Each data point represents one test data, with its x-axis and y-axis values corresponding to the RMSD (or DMAE) of React-OT and TS-DFM respectively.
Generally, the majority of data points in Fig. \ref{fig:rgd1_compare}a-f are located under black dashed lines, which is in accordance with the results of Tab. \ref{tab:res_rgd1}.
To our astonishment, there are several test data whose prediction accuracy of TS-DFM and React-OT differ greatly.
Figure \ref{fig:rgd1_compare}g presents three representative reactions that TS-DFM underperform React-OT, illustrating typical failure modes that lead to large structural errors.
In these cases, TS-DFM produces TSs with inaccurate torsion angles, inaccurate orientation of dissociated substructures, incorrect bond breakage/formation, and even incorrect reaction mechanisms.
Nevertheless, TS-DFM captures partially correct reaction mechanisms even in several extreme cases, such as identifying correct bond changes in the first two cases of Fig. \ref{fig:rgd1_compare}g.
In practice, these could provide reasonable initial guesses for subsequent structure optimization.
More test cases are provided in supplementary Fig. \ref{supp_fig:rgd1_reactot}, \ref{supp_fig:rgd1_comparable} and \ref{supp_fig:rgd1_tsdfm}.
It should be emphasized that these cases are not indicative of the overall performance, as TS-DFM maintains a higher average accuracy than React-OT.

\section{Discussion}
In this study, we propose TS-DFM, a novel distance-geometry-based flow matching framework for generating TSs with provided reactant and product structures. 
Operating in distance geometry space enables more effective capture of the fundamental nature of chemical reactions, i.e., the evolution of interatomic distances.
Consequently, TS-DFM not only achieves higher accuracy in predicting TSs than SOTA methods, but also demonstrates superior generalization ability on unseen reactions.
Even in some of the worst cases of TS-DFM, the predicted TS can still demonstrate partially correct reaction mechanisms.

Furthermore, TS-DFM holds great potential for integration into diverse workflows across various applications.
For example, the TSs predicted by TS-DFM provide better initial structures for TS search algorithms.
In our experiments, they are used as initial configurations for CI-NEB optimization with MLIP, improving both the convergence speed and the accuracy of converged structures.
This indicates that TS-DFM could offer substantial computational savings if quantum chemical accuracy is required for reaction pathway analysis.
As a proof of concept, we also demonstrate the application of TS-DFM in a workflow designed to discover alternative reaction pathways: by employing normal mode sampling to generate thermally activated reactants and products, followed by MLIP-based CI-NEB optimization, multiple TSs with different reaction mechanisms are identified in some chemical reactions.
Notably, in one specific case, we identify a reaction pathway that is more favorable than the reference in the dataset.
Such workflows hold significant promise for elucidating competitive reaction mechanisms, expanding reaction databases, and accelerating reaction discovery.

Currently, TS-DFM is limited to predicting the TS of uncatalyzed organic reactions.
Given that catalysts play an indispensable role in organic synthesis, modeling their influence is essential to further advance the framework's capabilities.
Beyond organic chemistry, extending TS-DFM to predict TSs in other domains like biology (e.g., protein dynamics and enzymatic catalysis) and materials science (e.g., surface chemistry and solid-state diffusion) also represents a promising research avenue.

\section{Methods}\label{sec11}

This section introduces the general framework, network design, evaluation metrics, and some implementation details of our proposed TS-DFM.
The following content of this section can be summarized as follows.
\begin{enumerate}
    \renewcommand{\labelenumi}{(\theenumi)}
    \item We elaborate on the theoretical foundations of flow matching as a generative model, and the formulation of TS-DFM for TS prediction task through a probabilistic generative modeling defined within a flow matching framework in molecular distance geometry space.
    \item  We describe the detailed design of TSDVNet, responsible for predicting the time-dependent velocity field of our TS-DFM framework. 
    We will introduce the key network module design for mapping the distance geometry of reactant, product and initial guessed TS to the velocity field.
    \item We provide the mathematical definitions of the metrics and explain their roles in assessing the accuracy of time-series predictions.
    \item We specify the training configurations for both TS-DFM and MLIP, and list the software packages and hyperparameters for our performed calculations.
\end{enumerate}

\subsection{Details about TS-DFM}
\subsubsection{Preliminaries of Flow Matching}
Let $x_0$ and $x_1$ be data points drawn from a source distribution $q_0$ and a target distribution $q_1$ in $\mathbb{R}^d$, respectively. 
A flow model \cite{lipman2024flowmatchingguidecode} defines a continuous-time Markov process ${x}_{t \in [0,1]}$ via a time-dependent flow $\psi_t$ applied to $x_0$, where $x_t = \psi_t(x_0)$ with $x_0 \sim q_0$. 
The goal of generative flow modeling is to learn a flow $\psi_t$ such that $x_1 = \psi_1(x_0)$ follows the target distribution $q_1$. 
The flow $\psi_t$ can be defined by a time-dependent velocity field $u_t : \mathbb{R}^d \to \mathbb{R}^d$ through an ordinary differential equation (ODE): $\frac{d}{dt} \psi_t(x) = u_t(\psi_t(x))$, with initial condition $\psi_0(x) = x$. 
A vector field $u_t$ is said to generate a probability path $p_t$ if its associated flow $\psi_t$ satisfies $x_t = \psi_t(x_0) \sim p_t$ for all $t \in [0,1]$.

Early work Continuous Normalizing Flow (CNF) \cite{NEURIPS2018_69386f6b} utilizes black-box ODE solvers to simulate the solution for training a parameterized flow model $\psi_t$.
Due to the computational burden and the gradient vanishing or exploding issues of numerical ODE simulations, CNFs are difficult to train.
To address these problems, Flow Matching (FM) \cite{lipman2023flow} offers a simulation-free approach by regressing a neural network $v_{\boldsymbol{\theta}}(x,t)$ to a target velocity field $u_t(x)$.
Since $u_t(x)$ is generally intractable, a common practice is to employ the Conditional Flow Matching (CFM) loss \cite{lipman2023flow, tong2024improving}, which is guaranteed to yield the same gradients with the FM objective:
\begin{equation*}
    \mathcal{L}_{\text{CFM}}(\boldsymbol{\theta}) = \mathbb{E}_{t\sim[0,1],x\sim p_t(x|z),z\sim q(z)}\|v_{\boldsymbol{\theta}}(x,t)-u_t(x|z)\|_2^2,
\end{equation*}
where $z$ denotes any condition independent of $x$, and $u_t(x|z)$ is a conditional vector field that generates $p_t(x|z)$ from $p_0(x|z)$.

There exist different ways to define $p_t(x|z)$ and $u_t(x|z)$. 
Our work is inspired by OT-CFM~\cite{tong2024improving} that utilizes the Gaussian conditional optimal transport path to define $p_t(x|z)$ and $u_t(x|z)$.
OT-CFM sets
\begin{equation*}
    \begin{aligned}
        z=(x_0, x_1) ~&\text{and}~q(z)=\pi(x_0, x_1), \\
        u_t(x|z)=x_1-x_0 ~&\text{and}~ p_t(x|z)=\mathcal{N}(x|t\cdot x_1 + (1-t)\cdot x_0, \sigma^2),
    \end{aligned}
\end{equation*}
where $\pi$ is the optimal coupling between distributions $q_0$ and $q_1$ under the squared Euclidean cost, i.e.,  
\begin{equation*}
    \pi = \arg\inf_{\pi'\in \Pi}\int c(x_0, x_1)\pi'(dx_0, dx_1),
\end{equation*}
where $\Pi$ is the set of couplings and $c(x_0, x_1)=\|x_0-x_1\|_2^2$ is the squared Euclidean distance. $u_t(x|z)=x_1-x_0$ stems from the optimal transport trajectory under the squared Euclidean cost, which is a straight line with constant velocity.
For this choice of $z$, $p_t(x|z)$ and $u_t(x|z)$, the corresponding marginal distribution $p_t(x)$ has the following boundary conditions
\begin{equation*}
    \begin{aligned}
    p_0(x)&=(q_0(\cdot)\ast \mathcal{N}(\cdot|0,\sigma^2))(x), \\
    p_1(x)&=(q_1(\cdot)\ast \mathcal{N}(\cdot|0,\sigma^2))(x),
    \end{aligned}
\end{equation*}
where $\ast$ denotes convolution operator.
As $\sigma\rightarrow 0$, the marginal boundary probabilities approach $q_0$ and $q_1$ respectively.

\subsubsection{Flow Matching on Distance Geometry for TS Prediction}
The dynamic evolution of interatomic distances between atom pairs is crucial for describing chemical reactions.
However, previous state-of-the-art methods such as OA-ReactDiff and React-OT operate in Cartesian coordinate space to predict TSs, which obscures this evolution process.
This potentially undermines the quality of their generated TSs.
Therefore, our proposed TS-DFM explicitly operates on the space of distance geometry to generate pairwise distance matrix of TS, which improves the accuracy of the predicted TS.

In distance geometry space, the task of double-ended TS prediction can be formulated as generating the pairwise distance matrix of TS $D_{\text{TS}}$ given the pairwise distance matrices of reactant $D_{\text{R}}$ and product $D_{\text{P}}$, along with the atom type vector $Z=[z_1, \dots,z_n],z_i\in\{1,\dots,118\}$.
Traditional TS optimization algorithms, such as NEB and Hessian-based methods, solve this problem by iteratively refining an initial guess of TS geometry towards the saddle point. 
From a probabilistic perspective, this process can be abstracted mathematically as transforming a distribution of initial guesses to a target distribution of true TSs.

Formally, let $\mathbb{P}(D_{\text{R}}, D_{\text{P}}, Z)$ denote the probability distribution over chemical reactions, characterized by $D_{\text{R}}$, $D_{\text{P}}$, and $Z$.
For any reaction sampled from $\mathbb{P}(D_{\text{R}}, D_{\text{P}}, Z)$, the algorithm operates on a pair of structures:
(1) The initial guessed distance geometry of TS, denoted as $D_{\text{TS},0}$, drawn from a conditional distribution $q_0(D_{\text{TS},0} \mid D_{\text{R}}, D_{\text{P}}, Z)$;
(2) The true TS distance geometry $D_{\text{TS},1}$, modeled by a conditional distribution $q_1(D_{\text{TS},1} \mid D_{\text{R}}, D_{\text{P}}, Z, D_{\text{TS},0})=\delta(D_{\text{TS},1}-f(D_{\text{R}}, D_{\text{P}}, Z, D_{\text{TS},0}))$, where $\delta$ denotes the Dirac delta distribution and $f$ represents the optimization algorithm. 

However, the optimization algorithm $f$ requires accurate PES, which relies on computationally prohibitive quantum chemistry calculations.
To circumvent this bottleneck, we propose using the Optimal Transport–Conditional Flow Matching (OT-CFM) framework to learn a data-driven surrogate as a substitution for the transport process guided by the PES.
In this approach, the distributions of initial guessed TSs and true TSs are treated as the source and target distributions, respectively, and a velocity field is learned to transport samples from the source to the target.

To utilize OT-CFM for generating TS, we first define the optimal coupling $\pi$ between the source and target distribution.
To maintain chemical validity, both $D_{\text{TS},0}$ and $D_{\text{TS},1}$ must correspond to the same reaction context $(D_{\text{R}}, D_{\text{P}}, Z)$. 
Therefore, we define the following joint distribution $\pi$ for this constraint:
\begin{equation*}
\begin{aligned}
    &\pi(D_{\text{R}}, D_{\text{P}}, Z, D_{\text{TS},0}, D_{\text{TS},1}) \\
    = &\mathbb{P}(D_{\text{R}}, D_{\text{P}}, Z) \cdot q_0(D_{\text{TS},0} | D_{\text{R}}, D_{\text{P}}, Z) \cdot q_1(D_{\text{TS},1} | D_{\text{R}}, D_{\text{P}}, Z, D_{\text{TS},0}).
\end{aligned}
\end{equation*}
Under this coupling $\pi$, any sample $(D_{\text{R}}^i, D_{\text{P}}^i, Z^i, D_{\text{TS},0}^i, D_{\text{TS},1}^i)$ guarantees that $D_{\text{TS},0}^i$ and $D_{\text{TS},1}^i$ are paired through the same reaction context $(D_{\text{R}}^i, D_{\text{P}}^i, Z^i)$.
Thus, the iterative refinement in traditional algorithms can be viewed as a deterministic transport trajectory from $D_{\text{TS},0}^i$ to $D_{\text{TS},1}^i$, guided by the local PES defined by $(D_{\text{R}}^i, D_{\text{P}}^i, Z^i)$.

Then we construct the source and target distributions $q_0(D_{\text{TS},0}, D_{\text{R}}, D_{\text{P}}, Z)$ and $q_1(D_{\text{TS},1}, D_{\text{R}}, D_{\text{P}}, Z)$ representing the distributions of initial guessed and true TS over chemical reactions respectively:
\begin{equation*}
    \begin{aligned}
    q_0(D_{\text{TS},0}, D_{\text{R}}, D_{\text{P}}, Z) &= \mathbb{P}(D_{\text{R}}, D_{\text{P}}, Z) \cdot q_0(D_{\text{TS},0}| D_{\text{R}}, D_{\text{P}}, Z), \\
    q_1(D_{\text{TS},1},D_{\text{R}},D_{\text{P}},  Z)&=\mathbb{P}(D_{\text{R}}, D_{\text{P}}, Z) \cdot q_1(D_{\text{TS},1}| D_{\text{R}}, D_{\text{P}}, Z).
    \end{aligned}
\end{equation*}
For the source distribution $q_0$, since it is typically chosen as an easy-to-sample distribution and is not required to be Gaussian in OT-CFM \cite{tong2024improving}, we adopt a practical choice inspired by the IDPP interpolation method \cite{10.1063/1.4878664}:
\begin{equation*}
    q_0(D_{\text{TS},0}| D_{\text{R}}, D_{\text{P}}, Z) = \delta\left(D_{\text{TS},0}-\frac{D_{\text{R}} + D_{\text{P}}}{2}\right).
\end{equation*}
This choice reduces the discrepancy between source and target distribution compared to generally utilized Gaussian noise as source distribution in flow matching model, and provides chemically more plausible initial structures than Cartesian linear interpolation used in React-OT and PSI-based model.
For $q_1(D_{\text{TS},1}| D_{\text{R}}, D_{\text{P}}, Z)$, we assume the following approximation:
\begin{equation*}
    q_1(D_{\text{TS},1}| D_{\text{R}}, D_{\text{P}}, Z)\approx q_1(D_{\text{TS},1}| D_{\text{R}}, D_{\text{P}}, Z,D_{\text{TS},0}),
\end{equation*}
which allows modeling TS generation without explicit dependence on the initial guess $D_{\text{TS},0}$, though this simplification may not hold in complex reactions where different initial guesses lead to varied TSs.

The coupling $\pi$ between initial and optimized TSs from the same reaction is used without explicitly solving the optimal transport problem.
The time-dependent probability path $p_t$ and velocity field $u_t$ are derived from the Gaussian probability path minimizing the 2-Wasserstein distance:
\begin{equation*}
     \begin{aligned}
    p_t(D_{\text{TS},t}|Z,D_{\text{R}},D_{\text{P}},D_{\text{TS},1})&=\mathcal{N}(D_{\text{TS},t}|t\cdot D_{\text{TS},1}+(1-t)\cdot D_{\text{TS},0},\sigma^2), \\
    u_t(D_{\text{TS},t}|Z,D_\text{R},D_\text{P},D_{\text{TS},1})&=D_{\text{TS},1}-D_{\text{TS},0}.
     \end{aligned}
\end{equation*}

Finally, we can construct a parameterized model $v_{\boldsymbol{\theta}}(D_{\text{TS},t},t|Z,D_\text{R},D_\text{P})$ to approximate the velocity field $u_t$ by minimizing the loss function:
\begin{equation*}
        \mathcal{L}_{\text{TS-DFM}}(\boldsymbol{\theta}) = \mathbb{E}_{t, p_t}
        \|v_{\boldsymbol{\theta}}(D_{\text{TS},t},t,|Z,D_\text{R},D_\text{P})-(D_{\text{TS},1}-D_{\text{TS},0})\|_2^2.
\end{equation*}
With a learned velocity field $v_{\boldsymbol{\theta}}(D_{\text{TS},t},t|Z,D_\text{R},D_\text{P})$, predicted distance geometry of TS $\hat{D}_{\text{TS},1}$ is obtained by numerical ODE simulation starting from $D_{\text{TS},0}$.
The corresponding Cartesian coordinates of predicted TS $\hat{R}_{\text{TS}}$ is recovered by nonlinear optimization, which will be elaborated later.
We also provide pseudocodes for the training and inference of TS-DFM in Supplementary Alg. \ref{supp_alg:training} and \ref{supp_alg:inference} respectively.

\subsubsection{Reconstruct Molecular Structure from Distance Geometry}
In TS-DFM, a velocity field of molecular distance geometry is learned for generating the pairwise distance matrix of TS.
Given the atomic coordinates $\vec{R}=[\boldsymbol{r}_1, \dots, \boldsymbol{r}_n]\in \mathbb{R}^{n\times 3}$, the pairwise distance matrix $D$ can be calculated easily from $\vec{R}$, with each element given by $d_{ij}=\|\boldsymbol{r}_i-\boldsymbol{r}_j\|_2$.
On the other hand, reconstructing the atomic coordinates $\vec{R}$ from pairwise distance matrix $D$ is nontrivial, which requires to perform nonlinear optimization.

From an initial atomic coordinates $\vec{R}_0$, the nonlinear optimization aims to find the optimal coordinates $\vec{R}^*$ that minimizes
$$
\vec{R}^* = \arg\min_{\vec{R}} \sum_{ij} \omega_{ij}(\hat{d}_{ij}-\|\boldsymbol{r}_i-\boldsymbol{r}_j\|_2)^2.
$$
where $\hat{d}_{ij}$ and $\omega_{ij}$ denote the predicted pairwise distance and the weight coefficient respectively.
In our experiments, the weight coefficient is set as $\omega_{ij}=1/\hat{d}_{ij}^2$ to emphasize short distances, which are closely associated with molecular bonding structure. 
The LBFGS optimizer is utilized for solving the optimization problem.
The initial atomic coordinates $\vec{R}_0$ is generated by the Multi-Dimensional Scaling algorithm \cite{torgersonMultidimensionalScalingSimilarity1965} using $\hat{d}_{ij}$.

\subsection{Network Structure of TSDVNet}
The detailed network structure TSDVNet for constructing $v_{\boldsymbol{\theta}}(D_{\text{TS},t},t|Z,D_\text{R},D_\text{P})$ in the flow matching framework is illustrated in Fig. \ref{fig:overview&structure}b.
The model takes atom type $Z$, timestep $t$, and three pairwise distance matrices $D_\text{R}, D_\text{P}, D_{\text{TS},t}$ as input and then outputs the predicted velocity field.
As shown in Fig. \ref{fig:overview&structure}b, the network contains two branches with the same architecture but different parameters.
The upper branch aims to generate the representations of $Z$ and $D_{\text{TS},t}$, which are then used to predict the velocity field.
While the lower branch is used to generate the conditional representations, where the conditional inputs $D_\text{R}, D_\text{P}, Z$ corresponding to the reactant and product will be processed.
To further enhance the learned representations and combining the information of reactant, product and TS, we also employ feature fusion operations between the two branches.

Within each branch, both atom and pair representations will be computed.
The Atom Embedding and Pair Embedding blocks encode atom type $Z$, pairwise distance matrix $D$ and timestep $t$ into atom representations $X=[\boldsymbol{x}_i, \dots,\boldsymbol{x}_n]\in \mathbb{R}^{n\times d_x}$ and pair representations $P=[\boldsymbol{p}_{ij}]_{i,j=1}^n\in \mathbb{R}^{n\times n \times d_p}$, where $d_x$ and $d_p$ represents the dimension of atom and pair representations respectively.
Then, these representations will be processed by $L$ update blocks.
In the $l$-th update block, the atom representations will be first updated by Cross-Molecule Message Passing block, fusing the information of $X_\text{R}^{(l)}$, $X_\text{P}^{(l)}$ and $X_{\text{TS},t}^{(l)}$.
Then, the Inner-Molecule Message Passing block updates the atom representations using the neighboring information within each molecule.
Utilizing the updated atom representations, the pair features will be updated by Mixing Update and Triangular Update blocks, focusing on local and global pair information respectively.
Finally, the Pairwise Distance Flow Prediction block utilizes the atom and pair representations $X_{\text{TS},t}$, $P_{\text{TS},t}$ to predict the velocity field $\boldsymbol{v}$.
In the following content of this section, each module will be introduced in detail.

\subsubsection{Atom and Pair Embedding}
The Atom Embedding block encodes the atom type $Z$, the neighboring information $j\in \mathcal{N}(i)$, and flow timestep $t$ into feature vectors $X$, which is defined as
\begin{equation*}
    \begin{aligned}
    \boldsymbol{x}_i &= \texttt{Embedding}(Z_i); \\
    \boldsymbol{x}_i &\leftarrow \boldsymbol{x}_i + \texttt{Time\_Embedding}(t); \\
    \boldsymbol{x}_i &\leftarrow W_N \left[\boldsymbol{x}_i,\sum_{j\in \mathcal{N}(i)}\texttt{Embedding}(\boldsymbol{x}_j)\odot \left(W_F\cdot \texttt{RBF}(d_{ij})\right)\right]+\boldsymbol{b}_N,
    \end{aligned}
\end{equation*}
where $[\dots]$ denotes concatenation operation.
Unless stated otherwise, $W_{\square}$ and $\boldsymbol{b}_{\square}$ are used to represent learnable parameters.
Within this block, the $\texttt{Embedding}$ function maps the atom types into embedded vectors,
the $\texttt{Time\_Embedding}$ function encodes the timestep $t$ into a vector by
\begin{equation*}
\begin{aligned}
&\boldsymbol{m}_t=t*\exp\left(-\frac{\log(10000)}{d/2-1}*[0,1,\dots,\frac{d}{2}-1]\right); \\
&\texttt{output} = [\sin(\boldsymbol{m}_t), \cos(\boldsymbol{m}_t)],\\
\end{aligned}
\end{equation*}
and the $\texttt{RBF}$ function encodes the atomic distances $d_{ij}$ into vectors with
\begin{equation*}
    e_{\text{RBF}}(d_{ij}) = \phi(d_{ij})\exp(-\beta_k(\exp(-d_{ij})-\mu_k)^2), \\
\end{equation*}
where
\begin{equation*}
\phi(d_{ij})=
\begin{cases}
\frac{1}{2}\left(\cos\left(\frac{\pi d_{ij}}{d_\mathrm{cut}}\right)+1\right), & \mathrm{if}~d_{ij}\leq d_\mathrm{cut} \\
0, & \mathrm{if}~d_{ij}>d_\mathrm{cut} 
\end{cases},
\end{equation*}
$\beta_k$ and $\mu_k$ are trainable parameters initialized by the settings in \cite{unkePhysNetNeuralNetwork2019}, and $d_{\text{cut}}$ is a hyperparameter representing the cutoff distance.
Based on the embedded atom features $X$, the Edge Embedding block generates pair representations $P$ by
\begin{equation*}
\begin{aligned}
    \boldsymbol{p}_{ij} = W_E[W_{S}\boldsymbol{x}_i+\boldsymbol{b}_{S},W_{D}\boldsymbol{x}_j+\boldsymbol{b}_{D},W_re_{\text{RBF}}(d_{ij})+\boldsymbol{b}_r]+\boldsymbol{b}_E.
\end{aligned}
\end{equation*}

\subsubsection{Update of Atom Representation}
At the $l$-th update block, the Cross Molecule Message Passing block and Inner Molecule Message Passing block are used to update atom representations.
Taking atom representation $\boldsymbol{x}_{i,\text{TS},t}$ (the $i$-th atom representation in $X_{\text{TS},t}$) as example, the Cross-Molecule Message Passing block in the upper branch of the network shown in Fig. \ref{fig:overview&structure}b updates $\boldsymbol{x}_{i,\text{TS},t}$ by fusing the cross-molecule information as follows
\begin{equation*}
\begin{aligned}
    \boldsymbol{x}_{i,\text{TS},t} \leftarrow \texttt{LayerNorm}(\boldsymbol{x}_{i,\text{TS},t}&+\sum_{j\in \mathcal{N}(i,\text{TS})}\texttt{Attn}(\boldsymbol{x}_{i,\text{R}},\boldsymbol{x}_{j,\text{TS},t},\boldsymbol{p}_{ij,\text{TS},t},d_{ij,\text{TS},t}) \\
    &+\sum_{j\in \mathcal{N}(i,\text{TS})}\texttt{Attn}(\boldsymbol{x}_{i,\text{P}},\boldsymbol{x}_{j,\text{TS},t},\boldsymbol{p}_{ij,\text{TS},t},d_{ij,\text{TS},t})),
\end{aligned}
\end{equation*}
where $j\in \mathcal{N}(i,\text{TS})$ denotes the neighbors of $i$-th atom determined by $D_{\text{TS},t}$. 
The atom representations $\boldsymbol{x}_{i,\text{R}}$ and $\boldsymbol{x}_{i,\text{P}}$ are updated in the same manner, using the corresponding block in the lower branch.
Then, the Inner-Molecule Message Passing blocks update atom representation $\boldsymbol{x}_i$ by utilizing the same $\texttt{Attn}$ function to implement inner-molecule message passing as
\begin{equation*}
\begin{aligned}
    \boldsymbol{x}_{i}\leftarrow \texttt{LayerNorm}(\boldsymbol{x}_{i}+\sum_{j\in \mathcal{N}(i)}\texttt{Attn}(\boldsymbol{x}_{i},\boldsymbol{x}_{j},\boldsymbol{p}_{ij},d_{ij})).
\end{aligned}
\end{equation*}
The $\texttt{Attn}$ function is used to compute interactions between pairs of atoms through the modified attention mechanism.
Take $\texttt{Attn}(\boldsymbol{x}_{i},\boldsymbol{x}_{j},\boldsymbol{p}_{ij},d_{ij}))$ as example, its output is calculated by 
\begin{equation*}
\begin{aligned}
    Q&=W_Q \boldsymbol{x}_{i}+\boldsymbol{b}_Q; \\
K&=W_K \boldsymbol{x}_{j}+\boldsymbol{b}_K; \\
D_K&=W_{D_K}e_{RBF}(d_{ij})+\boldsymbol{b}_{D_K}; \\
V&=W_V \boldsymbol{x}_{j}+\boldsymbol{b}_V; \\
D_V&=W_{D_V}e_{RBF}(d_{ij})+\boldsymbol{b}_{D_V}; \\
D_P&=W_P \boldsymbol{p}_{ij}+\boldsymbol{b}_P; \\
\texttt{output}&=V\odot D_V\odot\texttt{SiLU}(Q\odot K\odot D_K+D_P)\cdot \phi(d_{ij}),
\end{aligned}
\end{equation*}
where $\odot$ represents the Hadamard product of two matrices.

\subsubsection{Update of Edge Representation}
After updating the atom representations, the Edge Update module first updates the pair representation $\boldsymbol{p}_{ij}$ using its connected atom representations $\boldsymbol{x}_i, \boldsymbol{x}_j$, which is defined as
\begin{equation*}
\begin{aligned}
    \boldsymbol{x}_{map}&=W_{map}[\boldsymbol{x}_i, \boldsymbol{x}_j]+\boldsymbol{b}_{map}; \\
\boldsymbol{p}_a&=W_a[\boldsymbol{x}_{map},\boldsymbol{p}_{ij}]+\boldsymbol{b}_{a}; \\
\boldsymbol{p}_b&=W_b[\boldsymbol{x}_{map},\boldsymbol{p}_{ij}]+\boldsymbol{b}_{b}; \\
\Delta \boldsymbol{p}_{ij} &= W_{ab}\cdot\texttt{Flatten}(\boldsymbol{p}_a \otimes \boldsymbol{p}_b)+\boldsymbol{b}_{ab};\\
\boldsymbol{p}_{ij} &\leftarrow \texttt{LayerNorm}(\boldsymbol{p}_{ij}+\Delta \boldsymbol{p}_{ij}),
\end{aligned}
\end{equation*}
where $\otimes$ denotes the Kronecker product of two vectors.
After that, the Edge Triangular Update aggregates the information that forms triangular connections with atom $i,j$ to further enhance the pair representations, which is adopted from the implementation in AlphaFold2 \cite{jumperHighlyAccurateProtein2021a} and computed by
\begin{equation*}
\begin{aligned}
    \boldsymbol{p}_{ij,m}&=\texttt{Sigmoid}(W_{m_1}\boldsymbol{p}_{ij}+\boldsymbol{b}_{m_1})\odot(W_{m_2}\boldsymbol{p}_{ij}+\boldsymbol{b}_{m_2}); \\
\boldsymbol{p}_{ij,n}&=\texttt{Sigmoid}(W_{n_1}\boldsymbol{p}_{ij}+\boldsymbol{b}_{m_1})\odot(W_{n_2}\boldsymbol{p}_{ij}+\boldsymbol{b}_{m_2}) ;\\
\boldsymbol{o}_{ij} &= \texttt{LayerNorm}(\sum_{k} \boldsymbol{p}_{ik,m}\odot \boldsymbol{p}_{jk,n} + \sum_{k} \boldsymbol{p}_{ki,m}\odot \boldsymbol{p}_{kj,n}); \\
\boldsymbol{p}_{ij} &\leftarrow \texttt{LayerNorm}(\boldsymbol{p}_{ij}+\texttt{Sigmoid}(W_{o_1}\boldsymbol{p}_{ij}+\boldsymbol{b}_{o_1})\odot(W_{o_2}\boldsymbol{o}_{ij}+\boldsymbol{b}_{o_2})).
\end{aligned}
\end{equation*}

\subsubsection{Flow Prediction}
Finally, the Pairwise Distance Flow Prediction module predicts the velocity field $\boldsymbol{v}_{\text{TS},t}$ by incorporating the atom and pair representations after $L$ update blocks, which is calculated by
\begin{equation*}
\begin{aligned}
    \boldsymbol{x}_{map}&=W_{map}[\boldsymbol{x}_i, \boldsymbol{x}_j]+\boldsymbol{b}_{map}; \\
\boldsymbol{p}_a&=W_a[\boldsymbol{x}_{map},\boldsymbol{p}_{ij}]+\boldsymbol{b}_{a}; \\
\boldsymbol{p}_b&=W_b[\boldsymbol{x}_{map},\boldsymbol{p}_{ij}]+\boldsymbol{b}_{b}; \\
\boldsymbol{v}_{\text{TS},t} &= W_{ab}\cdot\texttt{Flatten}(\boldsymbol{p}_a \otimes \boldsymbol{p}_b)+\boldsymbol{b}_{ab}.\\
\end{aligned}
\end{equation*}

With the aforementioned network design, the velocity field $\boldsymbol{v}_{\text{TS},t}$ exhibits permutation equivariance.
More importantly, the dependence on reaction direction, i.e., the exchange of reactant and product, will also be eliminated.

\subsection{Evaluation Metrics}
As mentioned above, the TSs are located at the saddle points on the PES.
In evaluating the quality of predicted TSs, both their structural similarity to the reference TS and the properties associated with saddle points should be considered.
Therefore, the evaluation metrics used in our work are classified into two categories.
One category aims to evaluate the structural accuracy, including the following metrics:
\begin{enumerate}
    \renewcommand{\labelenumi}{(\theenumi)}
    \item Root Mean Squared Distance (RMSD):
    $$
    \text{RMSD} = \sqrt{\frac{1}{N}\sum_{i=1}^{N}\|\hat{\boldsymbol{x}}_i-\boldsymbol{x}_i\|_2^2},
    $$
    where $\hat{\boldsymbol{x}}_i$ and $\boldsymbol{x}_i$ denote the $i$-th atomic coordinates of predicted and true TS respectively, and $\hat{\boldsymbol{x}}_i$ is aligned w.r.t. $\boldsymbol{x}_i$ by Kabsch alignment algorithm \cite{Kabsch:a12999} before calculating RMSD.
    It mainly represents the global structural accuracy of the predicted TS.
    \item Distance Mean Absolute Error (DMAE):
    $$
    \text{DMAE} = \frac{1}{N(N-1)}\sum_{i\ne j}|\hat{d}_{ij}-d_{ij}|,
    $$
    where $d_{ij}=\|\boldsymbol{x}_i-\boldsymbol{x}_j\|_2$ and $\hat{d}_{ij}=\|\hat{\boldsymbol{x}}_i-\hat{\boldsymbol{x}}_j\|_2$.
    Compared with RMSD, DMAE focuses on local structural accuracy, making it a complementary metric for more nuanced structural accuracy evaluation.
\end{enumerate}
Another group of metrics are related to the potential energy properties of the predicted TS, which are derived from DFT calculations:
\begin{enumerate}
    \renewcommand{\labelenumi}{(\theenumi)}
    \item Potential energy error of TS $|\Delta E_\text{TS}|$:
    $$
    |\Delta E_\text{TS}|=|\hat{E}_\text{TS}-E_\text{TS}|,
    $$
    where $\hat{E}_\text{TS}$ and $E_\text{TS}$ denote the potential energy of the predicted and true TS.
    As the cases shown in \cite{https://doi.org/10.1002/wcms.70025}, a predicted TS with large $|\Delta E_\text{TS}|$ often exhibits an imperfect geometric structure that has not fully converged to saddle point, even with a low RMSD.
    Therefore, small $|\Delta E_\text{TS}|$ is a fundamental requirement for a valid result.

    \item The root mean square force norm of predicted TS $F_{\text{rms}}$:
    $$
    F_{\text{rms}}=\sqrt{\frac{1}{N}\sum_{i=1}^{N}\|\hat{\boldsymbol{f}}_i\|_2^2},
    $$
    where $\hat{\boldsymbol{f}}_i=\nabla_{\boldsymbol{\hat{x}_i}}\hat{E}_{\text{TS}}$.
    A low $F_{\text{rms}}$ could prove that the structure has converged to a stationary point.
    It serves as a prerequisite for further frequency analysis.
    
    \item The maximum force norm of the predicted TS $f_{\max}$:
    $$
    f_{\max}=\max_{i=1,\dots,N} \|\hat{\boldsymbol{f}}_i\|_2.
    $$
    Whereas $F_{\text{rms}}$ measures global convergence, $f_{\max}$ identifies localized geometric strain in a predicted TS, thus providing non-redundant information that supplements $F_{\text{rms}}$.
    
    \item \% of saddle points: The percentage of predicted TSs in the test set that have exactly one negative frequency—a key characteristic of saddle point.
    These frequencies are derived from harmonic analysis of the Hessian matrix.
    It directly indicates whether the predicted TS is located at the saddle point of PES or not.
    
\end{enumerate}
To summarize, the evaluation metrics described above assess the quality of predicted TSs from both structural and energetic perspectives. 
In comparison with previous works \cite{choiPredictionTransitionState2023a, duanAccurateTransitionState2023a, duanOptimalTransportGenerating2025a}, our study employs a more detailed set of metrics, thereby enabling a more comprehensive evaluation of different TS prediction methods.

\subsection{Implementation Details}
\subsubsection{Model Training}
\textbf{TS-DFM training}. 
The learnable velocity field network $v_{\boldsymbol{\theta}}$ is based on our proposed TSDVNet. 
This network comprises 6 update blocks, where both the atom and pair representations have a dimension of 128.
The cutoff distance is set as 20\AA.
We use mini-batch gradient descent algorithm with batch size of 32 to minimize $\mathcal{L}_{\text{TS-DFM}}$ using Adam optimizer \cite{DBLP:journals/corr/KingmaB14}, whose learning rate is initialized as 5e-4 and will be decayed by 0.8 if the performance is not improved for more than 40 epochs.
The noise scale $\sigma$ is set as 0.1 to improve the robustness of the learned velocity field.

\textbf{MLIP Training}. 
In our experiments on Transition1x dataset, a MLIP $E_{\boldsymbol{\theta}}(Z,\boldsymbol{X})$ is trained and then utilized as a surrogate PES for various tasks.
We utilize TorchMD-NET \cite{tholke2022equivariant}, an efficient and powerful equivariant graph neural network structure, to build the MLIP.
The number of message passing layers is set as 6, the hidden dimension is set as 256, and the cutoff distance is set as 10\AA.
The model parameter ${\boldsymbol{\theta}}$ is learned by optimizing the following loss function
\begin{equation*}
    \begin{aligned}
    \mathcal{L}_{\text{pot}}=\lambda(E_{\boldsymbol{\theta}}-E)^2+\frac{1-\lambda}{3N}\|\hat{\boldsymbol{F}}-\boldsymbol{F}\|^2,
    \end{aligned}
\end{equation*}
where $E$ denotes the true potential energy, $\hat{\boldsymbol{F}}=-\nabla_{\boldsymbol{X}}E_{\boldsymbol{\theta}}$ denotes the predicted atomic forces, and $\boldsymbol{F}$ represents the true atomic forces.
All the molecular configurations in the training and validation set are used for model training and validation, respectively.
We use mini-batch gradient descent algorithm with batch size of 50 to minimize the loss function using Adam optimizer \cite{DBLP:journals/corr/KingmaB14}.
The learning rate is initialized as 1e-3 and will be decayed by 0.75 if the performance is not improved for more than 25 epochs. 
The weight factor $\lambda$ in the loss function is set as 0.01 to put more emphasis on atomic forces, which is crucial for NEB calculation. 
The trained model achieves Mean Absolute Error (MAE) of 0.0696~eV for energy prediction and 0.0712~eV/\AA ~for atomic force prediction on the test set. 

\subsubsection{Calculation Details}
This section outlines the detailed implementation of the calculations performed in this study.
Additional information about these calculations is also provided in the Supplementary section \ref{supp_sec:tech_details}.

\textbf{NEB Calculation}. 
In this work, the NEB calculation is implemented by utilizing the interfaces of the Atomic Simulation Environment (ASE) \cite{hjorthlarsenAtomicSimulationEnvironment2017}.
Reactants and products are first relaxed for at most 500 steps using the BFGS optimizer with $\alpha=70$ and a maximum step size of 0.005\AA.
They are considered relaxed if the norm of the forces is lower than 0.01 eV/\AA.
Then, eleven images are used to represent the minimum energy path (MEP), whose initial configurations are generated by the IDPP \cite{10.1063/1.4878664} interpolation method.
If a predicted TS is provided, eight intermediate images are first placed along the reactant-TS and TS-product segments in a ratio determined by their respective DMAE values.
Each segment is then initialized using IDPP interpolation method, with reactant, TS, and product structures fixed as anchor points to constrain the initialization process.
After that, CI-NEB calculations are performed for at most 1000 steps, where the spring constant is set to 0.1 and the method is chosen as \texttt{aseneb}.
The MEP is found by the optimizer proposed in \cite{10.1063/1.5064465} with the default parameter.
During iteration, the path is considered converged if the maximal perpendicular force on the path is below 0.05 eV/\AA.

\textbf{Hessian Optimization}. 
We utilize Sella \cite{hermesSellaOpenSourceAutomationFriendly2022} for Hessian-based saddle point optimization.
Sella provides an automated algorithm that optimizes molecular structures to saddle points on the PES.
In our implementations, we utilize the default parameters provided by Sella.
The optimization runs for at most 1000 steps with a convergence threshold of 0.05 eV/\AA.

\textbf{IRC Calculation}.
The Intrinsic Reaction Coordinate (IRC) calculation traces the MEP on the PES, linking a TS to its reactant and product minima.
We utilize the \texttt{IRC} method provided in Sella for IRC calculations, where the parameters are set as default.
The calculation runs for a maximum of 1000 steps with a convergence threshold of 0.1 eV/\AA.
Then, the reactant and product structures will be further relaxed to ensure they are located at the local energy minima, following the same relaxation operations utilized in NEB calculations.

\textbf{Normal Mode Sampling}.
Normal mode sampling generates molecular configurations via harmonic approximation around a local minimum of PES by first calculating vibrational normal modes, then randomly displacing atoms along these modes with amplitudes scaled by the Boltzmann distribution at a specified temperature to reflect thermal fluctuations \cite{smithANI1DataSet2017a}.
In our implementations, the normal modes are obtained by DFT calculation, and the temperature is set as 300K.

\textbf{DFT Calculation}.
All DFT calculations are performed using PySCF \cite{10.1063/5.0006074}.
In performing calculations, the exchange-correlation functional and basis sets are consistent with the dataset.
We set the energy tolerance as 1e-6 in atomic units and the maximum number of SCF cycles as 200.

\section*{Data Availability}
The Transition1x dataset used in this work is publicly available at https://doi.org/10.6084/m9.figshare.19614657.v4.
The full RGD1 dataset is available at https://doi.org/10.6084/m9.figshare.21066901.v6.
The code for extracting subsets, together with the used subset of RGD1 in our experiments, will be available when the article is published.

\section*{Code Availability}
The code will be available as an open source repository on github when the article is published.

\bibliography{sn-bibliography}

@article{F29848000227,
  title = {Location of Transition States in Reaction Mechanisms},
  author = {Dewar, Michael J. S. and Healy, Eamonn F. and Stewart, James J. P.},
  year = {1984},
  journal = {Journal of the Chemical Society, Faraday Transactions 2: Molecular and Chemical Physics},
  volume = {80},
  number = {3},
  pages = {227--233},
  publisher = {The Royal Society of Chemistry},
  doi = {10.1039/F29848000227},
}

@article{eTransitionpathTheoryPathfinding2010,
  title = {Transition-Path Theory and Path-Finding Algorithms for the Study of Rare Events},
  author = {E, Weinan and {Vanden-Eijnden}, Eric},
  year = {2010},
  journal = {Annual Review of Physical Chemistry},
  volume = {61},
  number = {Volume 61, 2010},
  pages = {391--420},
  issn = {0066426X},
  doi = {10.1146/annurev.physchem.040808.090412},
}

@article{klucznikComputationalPredictionComplex2024,
  title = {Computational Prediction of Complex Cationic Rearrangement Outcomes},
  author = {Klucznik, Tomasz and Syntrivanis, Leonidas-Dimitrios and Ba{\'s}, Sebastian and {Mikulak-Klucznik}, Barbara and Moskal, Martyna and Szymku{\'c}, Sara and Mlynarski, Jacek and Gadina, Louis and Beker, Wiktor and Burke, Martin D. and Tiefenbacher, Konrad and Grzybowski, Bartosz A.},
  year = {2024},
  month = jan,
  journal = {Nature},
  volume = {625},
  number = {7995},
  pages = {508--515},
  issn = {1476-4687},
  doi = {10.1038/s41586-023-06854-3},
}

@article{zhangExploringFrontiersCondensedphase2024,
  title = {Exploring the Frontiers of Condensed-Phase Chemistry with a General Reactive Machine Learning Potential},
  author = {Zhang, Shuhao and Mako{\'s}, Ma{\l}gorzata Z. and Jadrich, Ryan B. and Kraka, Elfi and Barros, Kipton and Nebgen, Benjamin T. and Tretiak, Sergei and Isayev, Olexandr and Lubbers, Nicholas and Messerly, Richard A. and Smith, Justin S.},
  year = {2024},
  month = may,
  journal = {Nature Chemistry},
  volume = {16},
  number = {5},
  pages = {727--734},
  issn = {1755-4349},
  doi = {10.1038/s41557-023-01427-3},
}

@article{liuRehybridizationDynamicsPericyclic2023,
  title = {Rehybridization Dynamics into the Pericyclic Minimum of an Electrocyclic Reaction Imaged in Real-Time},
  author = {Liu, Y. and Sanchez, D. M. and Ware, M. R. and Champenois, E. G. and Yang, J. and Nunes, J. P. F. and Attar, A. and Centurion, M. and Cryan, J. P. and Forbes, R. and Hegazy, K. and Hoffmann, M. C. and Ji, F. and Lin, M.-F. and Luo, D. and Saha, S. K. and Shen, X. and Wang, X. J. and Mart{\'i}nez, T. J. and Wolf, T. J. A.},
  year = {2023},
  month = may,
  journal = {Nature Communications},
  volume = {14},
  number = {1},
  pages = {2795},
  issn = {2041-1723},
  doi = {10.1038/s41467-023-38513-6}
}

@article{10.1063/1.1691018,
  title = {A Growing String Method for Determining Transition States: {{Comparison}} to the Nudged Elastic Band and String Methods},
  author = {Peters, Baron and Heyden, Andreas and Bell, Alexis T. and Chakraborty, Arup},
  year = {2004},
  month = may,
  journal = {The Journal of Chemical Physics},
  volume = {120},
  number = {17},
  eprint = {https://pubs.aip.org/aip/jcp/article-pdf/120/17/7877/19133160/7877{\textbackslash}\_1{\textbackslash}\_online.pdf},
  pages = {7877--7886},
  issn = {0021-9606},
  doi = {10.1063/1.1691018},
}

@inbook{doi:10.1142/9789812839664_0016,
author = {Hannes Jónsson and Greg Mills and Karsten W. Jacobsen},
title = {Nudged elastic band method for finding minimum energy paths of transitions},
booktitle = {Classical and Quantum Dynamics in Condensed Phase Simulations},
year = {1998},
publisher = {World Scientific},
pages = {385-404},
doi = {10.1142/9789812839664_0016}
}

@article{Schreiner_2022,
  title = {{{NeuralNEB}}---Neural Networks Can Find Reaction Paths Fast},
  author = {Schreiner, Mathias and Bhowmik, Arghya and Vegge, Tejs and J{\o}rgensen, Peter Bj{\o}rn and Winther, Ole},
  year = {2022},
  month = dec,
  journal = {Machine Learning: Science and Technology},
  volume = {3},
  number = {4},
  pages = {045022},
  publisher = {IOP Publishing},
  doi = {10.1088/2632-2153/aca23e},
}

@article{yuanAnalyticalInitioHessian2024a,
  title = {Analytical Ab Initio Hessian from a Deep Learning Potential for Transition State Optimization},
  author = {Yuan, Eric C.-Y. and Kumar, Anup and Guan, Xingyi and Hermes, Eric D. and Rosen, Andrew S. and Z{\'a}dor, Judit and {Head-Gordon}, Teresa and Blau, Samuel M.},
  year = {2024},
  month = oct,
  journal = {Nature Communications},
  volume = {15},
  number = {1},
  pages = {8865},
  issn = {2041-1723},
  doi = {10.1038/s41467-024-52481-5},
}

@article{D0CP04670A,
  title = {Generating Transition States of Isomerization Reactions with Deep Learning},
  author = {Pattanaik, Lagnajit and Ingraham, John B. and Grambow, Colin A. and Green, William H.},
  year = {2020},
  journal = {Physical Chemistry Chemical Physics},
  volume = {22},
  number = {41},
  pages = {23618--23626},
  publisher = {The Royal Society of Chemistry},
  doi = {10.1039/D0CP04670A}
}

@article{D1SC01206A,
  title = {{{TSNet}}: Predicting Transition State Structures with Tensor Field Networks and Transfer Learning},
  author = {Jackson, Riley and Zhang, Wenyuan and Pearson, Jason},
  year = {2021},
  journal = {Chemical Science},
  volume = {12},
  number = {29},
  pages = {10022--10040},
  publisher = {The Royal Society of Chemistry},
  doi = {10.1039/D1SC01206A},
}

@article{choiPredictionTransitionState2023a,
  title = {Prediction of Transition State Structures of Gas-Phase Chemical Reactions via Machine Learning},
  author = {Choi, Sunghwan},
  year = {2023},
  month = mar,
  journal = {Nature Communications},
  volume = {14},
  number = {1},
  pages = {1168},
  issn = {2041-1723},
  doi = {10.1038/s41467-023-36823-3},
}

@article{10.1063/5.0055094,
  title = {Generative Adversarial Networks for Transition State Geometry Prediction},
  author = {Mako{\'s}, Ma{\l}gorzata Z. and Verma, Niraj and Larson, Eric C. and Freindorf, Marek and Kraka, Elfi},
  year = {2021},
  month = jul,
  journal = {The Journal of Chemical Physics},
  volume = {155},
  number = {2},
  pages = {024116},
  issn = {0021-9606},
  doi = {10.1063/5.0055094},
}

@article{duanAccurateTransitionState2023a,
  title = {Accurate Transition State Generation with an Object-Aware Equivariant Elementary Reaction Diffusion Model},
  author = {Duan, Chenru and Du, Yuanqi and Jia, Haojun and Kulik, Heather J.},
  year = {2023},
  month = dec,
  journal = {Nature Computational Science},
  volume = {3},
  number = {12},
  pages = {1045--1055},
  issn = {2662-8457},
  doi = {10.1038/s43588-023-00563-7}
}

@article{duanOptimalTransportGenerating2025a,
  title = {Optimal Transport for Generating Transition States in Chemical Reactions},
  author = {Duan, Chenru and Liu, Guan-Horng and Du, Yuanqi and Chen, Tianrong and Zhao, Qiyuan and Jia, Haojun and Gomes, Carla P. and Theodorou, Evangelos A. and Kulik, Heather J.},
  year = {2025},
  month = apr,
  journal = {Nature Machine Intelligence},
  volume = {7},
  number = {4},
  pages = {615--626},
  issn = {2522-5839},
  doi = {10.1038/s42256-025-01010-0}
}

@article{10.1063/1.4878664,
  title = {Improved Initial Guess for Minimum Energy Path Calculations},
  author = {Smidstrup, S{\o}ren and Pedersen, Andreas and Stokbro, Kurt and J{\'o}nsson, Hannes},
  year = {2014},
  month = jun,
  journal = {The Journal of Chemical Physics},
  volume = {140},
  number = {21},
  pages = {214106},
  issn = {0021-9606},
  doi = {10.1063/1.4878664},
}

@article{schreinerTransition1xDatasetBuilding2022a,
  title = {Transition1x - a Dataset for Building Generalizable Reactive Machine Learning Potentials},
  author = {Schreiner, Mathias and Bhowmik, Arghya and Vegge, Tejs and Busk, Jonas and Winther, Ole},
  year = {2022},
  month = dec,
  journal = {Scientific Data},
  volume = {9},
  number = {1},
  pages = {779},
  issn = {2052-4463},
  doi = {10.1038/s41597-022-01870-w},
}

@article{smithANI1DataSet2017a,
  title = {{{ANI-1}}, {{A}} Data Set of 20 Million Calculated off-Equilibrium Conformations for Organic Molecules},
  author = {Smith, Justin S. and Isayev, Olexandr and Roitberg, Adrian E.},
  year = {2017},
  month = dec,
  journal = {Scientific Data},
  volume = {4},
  number = {1},
  pages = {170193},
  issn = {2052-4463},
  doi = {10.1038/sdata.2017.193},
}

@article{PhysRevLett.108.058301,
  title = {Fast and Accurate Modeling of Molecular Atomization Energies with Machine Learning},
  author = {Rupp, Matthias and Tkatchenko, Alexandre and M\"uller, Klaus-Robert and von Lilienfeld, O. Anatole},
  journal = {Phys. Rev. Lett.},
  volume = {108},
  issue = {5},
  pages = {058301},
  numpages = {5},
  year = {2012},
  month = {Jan},
  publisher = {American Physical Society},
  doi = {10.1103/PhysRevLett.108.058301},
}

@article{zhaoComprehensiveExplorationGraphically2023a,
  title = {Comprehensive Exploration of Graphically Defined Reaction Spaces},
  author = {Zhao, Qiyuan and Vaddadi, Sai Mahit and Woulfe, Michael and Ogunfowora, Lawal A. and Garimella, Sanjay S. and Isayev, Olexandr and Savoie, Brett M.},
  year = {2023},
  month = mar,
  journal = {Scientific Data},
  volume = {10},
  number = {1},
  pages = {145},
  issn = {2052-4463},
  doi = {10.1038/s41597-023-02043-z}
}

@misc{lipman2024flowmatchingguidecode,
  title={Flow Matching Guide and Code}, 
  author={Yaron Lipman and Marton Havasi and Peter Holderrieth and Neta Shaul and Matt Le and Brian Karrer and Ricky T. Q. Chen and David Lopez-Paz and Heli Ben-Hamu and Itai Gat},
  year    = {2024},
  howpublished = {Preprint at \url{https://arxiv.org/abs/2412.06264}},
}

@inproceedings{NEURIPS2018_69386f6b,
  title = {Neural Ordinary Differential Equations},
  booktitle = {Advances in Neural Information Processing Systems},
  author = {Chen, Ricky T. Q. and Rubanova, Yulia and Bettencourt, Jesse and Duvenaud, David K},
  year = {2018},
  volume = {31},
  publisher = {Curran Associates, Inc.},
  editor = {Bengio, S. and Wallach, H. and Larochelle, H. and Grauman, K. and {Cesa-Bianchi}, N. and Garnett, R.},
}

@misc{lipman2023flow,
    title={Flow Matching for Generative Modeling},
    author={Yaron Lipman and Ricky T. Q. Chen and Heli Ben-Hamu and Maximilian Nickel and Matthew Le},
    howpublished ={In The Eleventh International Conference on Learning Representations},
    year={2023},
}

@article{jumperHighlyAccurateProtein2021a,
  title = {Highly Accurate Protein Structure Prediction with {{AlphaFold}}},
  author = {Jumper, John and Evans, Richard and Pritzel, Alexander and Green, Tim and Figurnov, Michael and Ronneberger, Olaf and Tunyasuvunakool, Kathryn and Bates, Russ and {\v Z}{\'i}dek, Augustin and Potapenko, Anna and Bridgland, Alex and Meyer, Clemens and Kohl, Simon A. A. and Ballard, Andrew J. and Cowie, Andrew and {Romera-Paredes}, Bernardino and Nikolov, Stanislav and Jain, Rishub and Adler, Jonas and Back, Trevor and Petersen, Stig and Reiman, David and Clancy, Ellen and Zielinski, Michal and Steinegger, Martin and Pacholska, Michalina and Berghammer, Tamas and Bodenstein, Sebastian and Silver, David and Vinyals, Oriol and Senior, Andrew W. and Kavukcuoglu, Koray and Kohli, Pushmeet and Hassabis, Demis},
  year = {2021},
  month = aug,
  journal = {Nature},
  volume = {596},
  number = {7873},
  pages = {583--589},
  issn = {1476-4687},
  doi = {10.1038/s41586-021-03819-2},
}

@misc{DBLP:journals/corr/KingmaB14,
  author       = {Diederik P. Kingma and
                  Jimmy Ba},
  title        = {Adam: {A} Method for Stochastic Optimization},
  howpublished    = {In 3rd International Conference on Learning Representations},
  year         = {2015},
}

@misc{tholke2022equivariant,
title={Equivariant Transformers for Neural Network based Molecular Potentials},
author={Philipp Th{\"o}lke and Gianni De Fabritiis},
howpublished ={In International Conference on Learning Representations},
year={2022},
}

@article{torgersonMultidimensionalScalingSimilarity1965,
  title = {Multidimensional Scaling of Similarity},
  author = {Torgerson, Warren S.},
  year = {1965},
  journal = {Psychometrika},
  volume = {30},
  number = {4},
  pages = {379--393},
  issn = {1860-0980},
  doi = {10.1007/BF02289530}
}

@article{https://doi.org/10.1002/wcms.1354,
author = {Dewyer, Amanda L. and Argüelles, Alonso J. and Zimmerman, Paul M.},
title = {Methods for exploring reaction space in molecular systems},
journal = {WIREs Computational Molecular Science},
volume = {8},
number = {2},
pages = {e1354},
doi = {https://doi.org/10.1002/wcms.1354},
year = {2018}
}

@article{durantePredictionOrganicReaction2000,
  title = {Prediction of {{Organic Reaction Products}}:\, {{Determining}} the {{Best Reaction Conditions}}},
  author = {Durante, Marco and Sello, Guido},
  year = {2000},
  month = mar,
  journal = {Journal of Chemical Information and Computer Sciences},
  volume = {40},
  number = {2},
  pages = {221--235},
  publisher = {American Chemical Society},
  issn = {0095-2338},
  doi = {10.1021/ci990430c},
}

@article{kwonComputationalTransitionStateDesign2018,
  title = {Computational {{Transition-State Design Provides Experimentally Verified Cr}}({{P}},{{N}}) {{Catalysts}} for {{Control}} of {{Ethylene Trimerization}} and {{Tetramerization}}},
  author = {Kwon, Doo-Hyun and Fuller, Jack T. III and Kilgore, Uriah J. and Sydora, Orson L. and Bischof, Steven M. and Ess, Daniel H.},
  year = {2018},
  month = feb,
  journal = {ACS Catalysis},
  volume = {8},
  number = {2},
  pages = {1138--1142},
  publisher = {American Chemical Society},
  doi = {10.1021/acscatal.7b04026}
}

@article{hjorthlarsenAtomicSimulationEnvironment2017,
  title = {The Atomic Simulation Environment---a {{Python}} Library for Working with Atoms},
  author = {Hjorth Larsen, Ask and J{\o}rgen Mortensen, Jens and Blomqvist, Jakob and Castelli, Ivano E and Christensen, Rune and Du{\l}ak, Marcin and Friis, Jesper and Groves, Michael N and Hammer, Bj{\o}rk and Hargus, Cory and Hermes, Eric D and Jennings, Paul C and Bjerre Jensen, Peter and Kermode, James and Kitchin, John R and Leonhard Kolsbjerg, Esben and Kubal, Joseph and Kaasbjerg, Kristen and Lysgaard, Steen and Bergmann Maronsson, J{\'o}n and Maxson, Tristan and Olsen, Thomas and Pastewka, Lars and Peterson, Andrew and Rostgaard, Carsten and Schi{\o}tz, Jakob and Sch{\"u}tt, Ole and Strange, Mikkel and Thygesen, Kristian S and Vegge, Tejs and Vilhelmsen, Lasse and Walter, Michael and Zeng, Zhenhua and Jacobsen, Karsten W},
  year = {2017},
  month = jun,
  journal = {Journal of Physics: Condensed Matter},
  volume = {29},
  number = {27},
  pages = {273002},
  publisher = {IOP Publishing},
  issn = {0953-8984},
  doi = {10.1088/1361-648X/aa680e},
}

@article{hermesSellaOpenSourceAutomationFriendly2022,
  title = {Sella, an {{Open-Source Automation-Friendly Molecular Saddle Point Optimizer}}},
  author = {Hermes, Eric D. and Sargsyan, Khachik and Najm, Habib N. and Z{\'a}dor, Judit},
  year = {2022},
  month = nov,
  journal = {Journal of Chemical Theory and Computation},
  volume = {18},
  number = {11},
  pages = {6974--6988},
  publisher = {American Chemical Society},
  issn = {1549-9618},
  doi = {10.1021/acs.jctc.2c00395},
}

@article{10.1063/1.5064465,
    author = {Makri, Stela and Ortner, Christoph and Kermode, James R.},
    title = {A preconditioning scheme for minimum energy path finding methods},
    journal = {The Journal of Chemical Physics},
    volume = {150},
    number = {9},
    pages = {094109},
    year = {2019},
    month = {03},
    issn = {0021-9606},
    doi = {10.1063/1.5064465},
}

@article{10.1063/5.0006074,
  title = {Recent Developments in the {{PySCF}} Program Package},
  author = {Sun, Qiming and Zhang, Xing and Banerjee, Samragni and Bao, Peng and Barbry, Marc and Blunt, Nick S. and Bogdanov, Nikolay A. and Booth, George H. and Chen, Jia and Cui, Zhi-Hao and Eriksen, Janus J. and Gao, Yang and Guo, Sheng and Hermann, Jan and Hermes, Matthew R. and Koh, Kevin and Koval, Peter and Lehtola, Susi and Li, Zhendong and Liu, Junzi and Mardirossian, Narbe and McClain, James D. and Motta, Mario and Mussard, Bastien and Pham, Hung Q. and Pulkin, Artem and Purwanto, Wirawan and Robinson, Paul J. and Ronca, Enrico and Sayfutyarova, Elvira R. and Scheurer, Maximilian and Schurkus, Henry F. and Smith, James E. T. and Sun, Chong and Sun, Shi-Ning and Upadhyay, Shiv and Wagner, Lucas K. and Wang, Xiao and White, Alec and Whitfield, James Daniel and Williamson, Mark J. and Wouters, Sebastian and Yang, Jun and Yu, Jason M. and Zhu, Tianyu and Berkelbach, Timothy C. and Sharma, Sandeep and Sokolov, Alexander Yu. and Chan, Garnet Kin-Lic},
  year = {2020},
  month = jul,
  journal = {The Journal of Chemical Physics},
  volume = {153},
  number = {2},
  pages = {024109},
  issn = {0021-9606},
  doi = {10.1063/5.0006074},
}

@article{tong2024improving,
title={Improving and generalizing flow-based generative models with minibatch optimal transport},
author={Alexander Tong and Kilian FATRAS and Nikolay Malkin and Guillaume Huguet and Yanlei Zhang and Jarrid Rector-Brooks and Guy Wolf and Yoshua Bengio},
journal={Transactions on Machine Learning Research},
issn={2835-8856},
year={2024},
}

@article{https://doi.org/10.1002/wcms.70025,
author = {Beaglehole, Isaac W. and Pemberton, Miles J. and Farrar, Elliot H. E. and Grayson, Matthew N.},
title = {Machine Learning Transition State Geometries and Applications in Reaction Property Prediction},
journal = {WIREs Computational Molecular Science},
volume = {15},
number = {3},
pages = {e70025},
doi = {https://doi.org/10.1002/wcms.70025},
note = {e70025 CMS-1187.R1},
year = {2025}
}

@article{gimarcShapesSimplePolyatomic1970,
  title = {Shapes of Simple Polyatomic Molecules and Ions.  {{I}}.  {{Series HAAH}} and {{BAAB}}},
  author = {Gimarc, Benjamin M.},
  year = {1970},
  month = jan,
  journal = {Journal of the American Chemical Society},
  volume = {92},
  number = {2},
  pages = {266--275},
  publisher = {American Chemical Society},
  issn = {0002-7863},
  doi = {10.1021/ja00705a007}
}

@article{tishchenkoUnifiedPerspectiveHydrogen2008,
  title = {A {{Unified Perspective}} on the {{Hydrogen Atom Transfer}} and {{Proton-Coupled Electron Transfer Mechanisms}} in {{Terms}} of {{Topographic Features}} of the {{Ground}} and {{Excited Potential Energy Surfaces As Exemplified}} by the {{Reaction}} between {{Phenol}} and {{Radicals}}},
  author = {Tishchenko, Oksana and Truhlar, Donald G. and Ceulemans, Arnout and Nguyen, Minh Tho},
  year = {2008},
  month = jun,
  journal = {Journal of the American Chemical Society},
  volume = {130},
  number = {22},
  pages = {7000--7010},
  publisher = {American Chemical Society},
  issn = {0002-7863},
  doi = {10.1021/ja7102907},
}

@article{denzelHessianMatrixUpdate2020,
  title = {Hessian {{Matrix Update Scheme}} for {{Transition State Search Based}} on {{Gaussian Process Regression}}},
  author = {Denzel, Alexander and K{\"a}stner, Johannes},
  year = {2020},
  month = aug,
  journal = {Journal of Chemical Theory and Computation},
  volume = {16},
  number = {8},
  pages = {5083--5089},
  publisher = {American Chemical Society},
  issn = {1549-9618},
}

@article{zengComplexReactionProcesses2020,
  title = {Complex Reaction Processes in Combustion Unraveled by Neural Network-Based Molecular Dynamics Simulation},
  author = {Zeng, Jinzhe and Cao, Liqun and Xu, Mingyuan and Zhu, Tong and Zhang, John Z. H.},
  year = {2020},
  month = nov,
  journal = {Nature Communications},
  volume = {11},
  number = {1},
  pages = {5713},
  issn = {2041-1723},
}

@article{zhaoSimultaneouslyImprovingReaction2021,
  title = {Simultaneously Improving Reaction Coverage and Computational Cost in Automated Reaction Prediction Tasks},
  author = {Zhao, Qiyuan and Savoie, Brett M.},
  year = {2021},
  month = jul,
  journal = {Nature Computational Science},
  volume = {1},
  number = {7},
  pages = {479--490},
  issn = {2662-8457},
}

@article{Kabsch:a12999,
author = "Kabsch, W.",
title = "{A solution for the best rotation to relate two sets of vectors}",
journal = "Acta Crystallographica Section A",
year = "1976",
volume = "32",
number = "5",
pages = "922--923",
month = "Sep",
}

@article{unkePhysNetNeuralNetwork2019,
  title = {{{PhysNet}}: {{A Neural Network}} for {{Predicting Energies}}, {{Forces}}, {{Dipole Moments}}, and {{Partial Charges}}},
  author = {Unke, Oliver T. and Meuwly, Markus},
  year = {2019},
  month = jun,
  journal = {Journal of Chemical Theory and Computation},
  volume = {15},
  number = {6},
  pages = {3678--3693},
  publisher = {American Chemical Society},
  issn = {1549-9618},
}

@article{10.1093/bioinformatics/btu829,
  title = {{{3Dmol}}.Js: Molecular Visualization with {{WebGL}}},
  author = {Rego, Nicholas and Koes, David},
  year = {2014},
  month = dec,
  journal = {Bioinformatics (Oxford, England)},
  volume = {31},
  number = {8},
  pages = {1322--1324},
  issn = {1367-4803},
}

@article{https://doi.org/10.1002/bkcs.10334,
    author = {Kim, Yeonjoon and Kim, Woo Youn},
    title = {Universal Structure Conversion Method for Organic Molecules: From Atomic Connectivity to Three-Dimensional Geometry},
    journal = {Bulletin of the Korean Chemical Society},
    volume = {36},
    number = {7},
    pages = {1769-1777},
    year = {2015}
}

\section*{Acknowledgements}
This work was supported by the National Key R\&D Program 2021YFA1003002, NSFC (12125104, 12426313).

\section*{Author contributions}
Y.L.: Conceptualization, methodology, software, writing of original draft and visualization.
X.G.: Methodology, review and editing.
J.S.: Methodology, review and editing, funding.

\end{document}


\title[Article Title]{\textit{Supplementary Information} for ``Generating transition states of chemical reactions via distance-geometry-based flow matching''}


\author[1]{\fnm{Yufei} \sur{Luo}}\email{yufei\_luo@stu.xjtu.edu.cn}

\author[1,3]{\fnm{Xiang} \sur{Gu}}\email{xianggu@xjtu.edu.cn}

\author*[1,2,3]{\fnm{Jian} \sur{Sun}}\email{jiansun@xjtu.edu.cn}

\affil*[1]{
\orgdiv{School of Mathematics and Statistics}, 
\orgname{Xi'an Jiaotong University}, 
\orgaddress{
\city{Xi'an}, 
\state{Shannxi Province}, 
\country{China}
}}
\affil*[2]{\orgdiv{State Industry-Education Integration Center for Medical Innovations at Xi'an Jiaotong University}, 
\orgaddress{
\city{Xi'an}, 
\state{Shannxi Province}, 
\country{China}
}}
\affil*[3]{\orgdiv{National Engineering Laboratory for Big Data Algorithms and Analysis Technologies}, 
\orgname{Xi'an Jiaotong University}, 
\orgaddress{
\city{Xi'an}, 
\state{Shannxi Province}, 
\country{China}
}}

\maketitle

\renewcommand{\thefigure}{S\arabic{figure}}
\setcounter{figure}{0} 
\renewcommand{\thetable}{S\arabic{table}}
\setcounter{table}{0} 
\renewcommand{\thealgorithm}{S\arabic{algorithm}}
\setcounter{algorithm}{0} 

\section{Abbreviation}
The following is a list of abbreviations utilized in the main article.
\begin{itemize}
    \item NEB: Nudged Elastic Band
    \item MEP: Minimum Energy Path
    \item PES: Potential Energy Surface
    \item CI-NEB: Climbing Image Nudged Elastic Band
    \item IRC: Intrinsic Reaction Coordinate
    \item TS: Transition State
    \item RMSD: Root Mean Squared Distance
    \item DMAE: Distance Mean Absolute Error
\end{itemize}

\section{Technical Details about Calculation Methods}
\label{supp_sec:tech_details}
\subsection{NEB Calculation}
NEB is a method for finding the MEP and TS given the reactant and product of a chemical reaction.
It begins by optimizing the geometries of the reactant and product to ensure they represent local minima on the PES.
A series of intermediate configurations are then created via interpolation between two endpoints, forming a chain connected by virtual spring forces that maintain uniform spacing along the path.
During optimization, the forces acting on each image are decomposed into components parallel and perpendicular to the path. 
The parallel component is eliminated to prevent images from sliding toward two endpoints, while the ``nudging'' force perpendicular to the path drives the images toward the MEP.
There is no guarantee that the image with the highest energy corresponds to the maximal energy of the MEP at convergence, i.e., the maximum might lie between two images.
To solve this problem, CI-NEB is often utilized to further maximize its energy by following the gradient on the PES parallel to the current path.
The MEP is considered converged once the maximal perpendicular force on the path is below a pre-defined threshold.

\subsection{Hessian Optimization}
For Hessian optimization calculations, we utilize Sella \cite{hermesSellaOpenSourceAutomationFriendly2022} as the optimizer, where an automated saddle point optimization algorithm is implemented.
It could automatically generate internal coordinates from Cartesian geometries, with the ability to handle problematic coordinates during iteration.
Instead of evaluating the full Hessian, it uses iterative diagonalization to identify the reaction coordinate and updates the approximate Hessian via multi-secant TS-BFGS Hessian update.
For constrained optimization to a saddle point, it combines null-space sequential quadratic programming (SQP) with restricted-step partitioned rational function optimization (RS-PRFO), splitting displacements into constraint-correction and optimization steps to move toward saddle points by maximizing energy along the reaction coordinate while minimizing in other directions.
Additionally, it employs geodesic stepping to realize displacements in redundant internal coordinates, accounting for the curvature of the coordinate manifold to ensure physically valid configurations and improve convergence efficiency.

\subsection{IRC Calculation}
The intrinsic reaction coordinate (IRC) calculation aims to trace the MEP on the PES, connecting a TS back to its corresponding reactant and product.
Starting precisely from the located TS geometry, the IRC path follows the direction of steepest descent in mass-weighted coordinates downhill towards lower energy. 
This involves numerical integration of the differential equations governing the path, typically using small steps along the negative gradient of the PES, while accounting for atomic masses to define the intrinsic reaction coordinate.
The calculation proceeds iteratively in forward and reverse directions from the TS until it reaches stable minima corresponding to the reactant and product states, thereby revealing the complete reaction mechanism.

\subsection{Normal Mode Sampling}
Normal mode sampling generates molecular configurations by exploiting the harmonic approximation around a structure located at the local minimum of PES \cite{smithANI1DataSet2017a}.
It first calculates the system's vibrational normal modes, i.e., the eigenvectors of the mass-weighted Hessian matrix, then randomly displaces atoms along these modes.
The amplitudes of random displacements are scaled by the Boltzmann distribution at a specified temperature, ensuring the sampled configurations reflect the correct quantum or classical thermal fluctuations for that temperature.

\section{Supplementary Tables}
\begin{table*}[!h]
  \centering
  \caption{\textbf{Results of ablation studies for TS-DFM.}
  In $\text{TS-DFM}^-$, the timestep is fixed at $t=0$ and one-step prediction is executed at inference.
  And $\text{TS-DFM}^{--}$ further removes linear interpolation on pairwise distance and uses interpolation on Cartesian coordinates instead to generate initial guessed TS.
  Network structure is the only difference between $\text{TS-DFM}^{--}$ and the PSI-based model.
  }
   \resizebox{\textwidth}{!}{
    \begin{tabular}{l|cccccccccccc}
    \toprule
          \multirow{2}[1]{*}{Approaches} & \multirow{2}[1]{*}{Parameters} & \multirow{2}[1]{*}{\makecell{\% of \\[-3pt] saddle points}} & \multicolumn{2}{c}{RMSD (\AA)} & \multicolumn{2}{c}{DMAE (\AA)} & \multicolumn{2}{c}{$|\Delta E_{\text{TS}}|$ (eV)} & \multicolumn{2}{c}{$f_{\max}$ (eV/\AA)} & \multicolumn{2}{c}{$F_\text{rms}$ (eV/\AA)} \\
           & & & mean & median & mean  & median & mean  & median & mean  & median & mean  & median \\
          \midrule
    TS-DFM$^{--}$ & 5.2M  & 31.80\% & 0.2820 & 0.2217 & 0.1024 & 0.0862 & 0.6894 & 0.3783 & 3.7043 & 3.1038 & 1.6990 & 1.4958 \\
    TS-DFM$^{-}$ & 5.2M  & 48.60\% & 0.2157 & 0.1427 & 0.0742 & 0.0565 & 0.3459 & 0.1545 & 2.3607 & 1.6893 & 1.1037 & 0.8977 \\
    TS-DFM  & 5.2M  & 65.50\% & 0.1828 & 0.0999 & 0.0607 & 0.0402 & 0.1614 & 0.0617 & 1.6482 & 1.2259 & 0.8227 & 0.6100 \\
    \bottomrule
    \end{tabular}
    }
    \label{supp_tab:ablation}
\end{table*}

\newpage

\section{Supplementary Figures}
\begin{figure}[ht]
    \centering
    \includegraphics[width=0.8\textwidth]{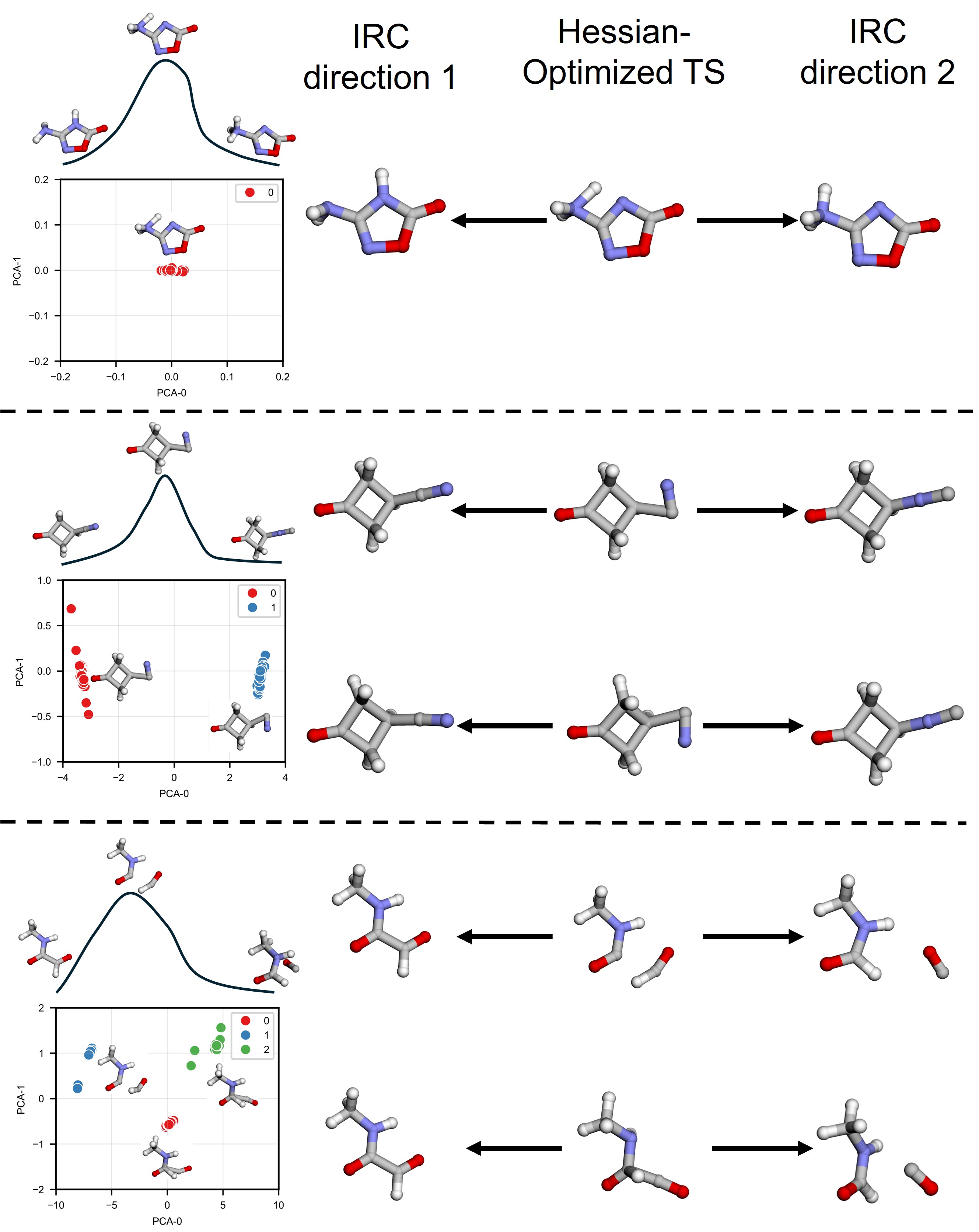}
    \caption{\textbf{IRC validation results for the generated samples in TS exploration}.
    The H, C, N and O atoms are colored as white, gray, blue and red respectively.
    In the bottom example, since the TS structure corresponds to cluster 0 has a Hessian-optimized structure identical to that of cluster 2, its IRC validation is omitted.}
    \label{supp_fig:rand_gen_irc}
\end{figure}

\begin{figure}[!ht]
    \centering
    \includegraphics[width=\textwidth]{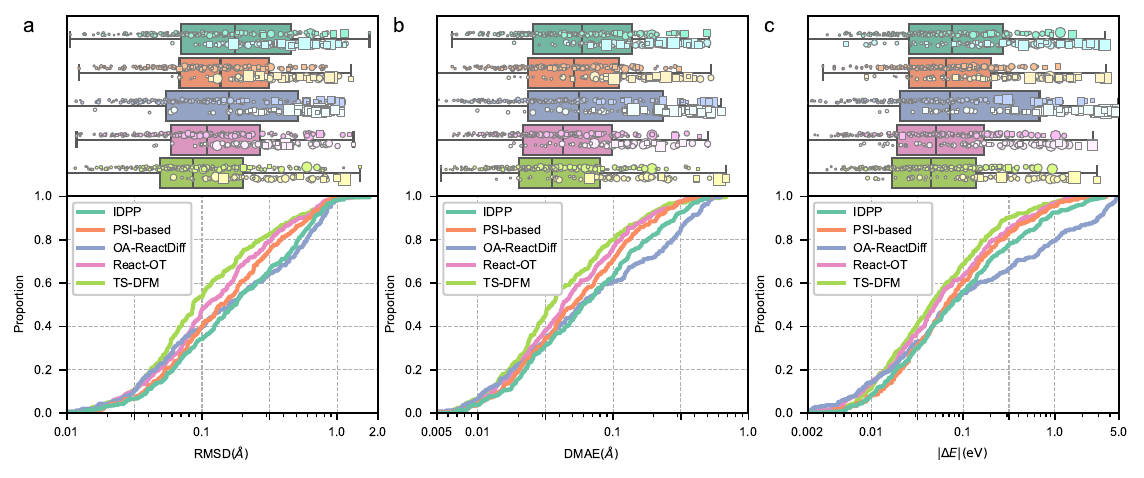}
    \caption{\textbf{Structural and energetic performance of different initialization methods in MLIP-based CI-NEB optimization}.
    \textbf{a}, \textbf{b} and \textbf{c}, The boxplots (upper) and cumulative probabilities (lower) for RMSD, DMAE and $|\Delta E_{\text{TS}}|$ of the predicted TS structures on test dataset respectively.
    The prediction error for each test data is also plotted as a point on the boxplot, where the point’s size is proportional to its $f_{\max}$ value, its color (dark or light) denotes whether the predicted TS structure lies at a saddle point or not, and its shape (a circle or square) denotes whether the NEB calculation converged or not.}
    \label{suppl_fig:res_ts1x_neb}
\end{figure}

\begin{figure}[ht]
    \centering
    \includegraphics[width=\textwidth]{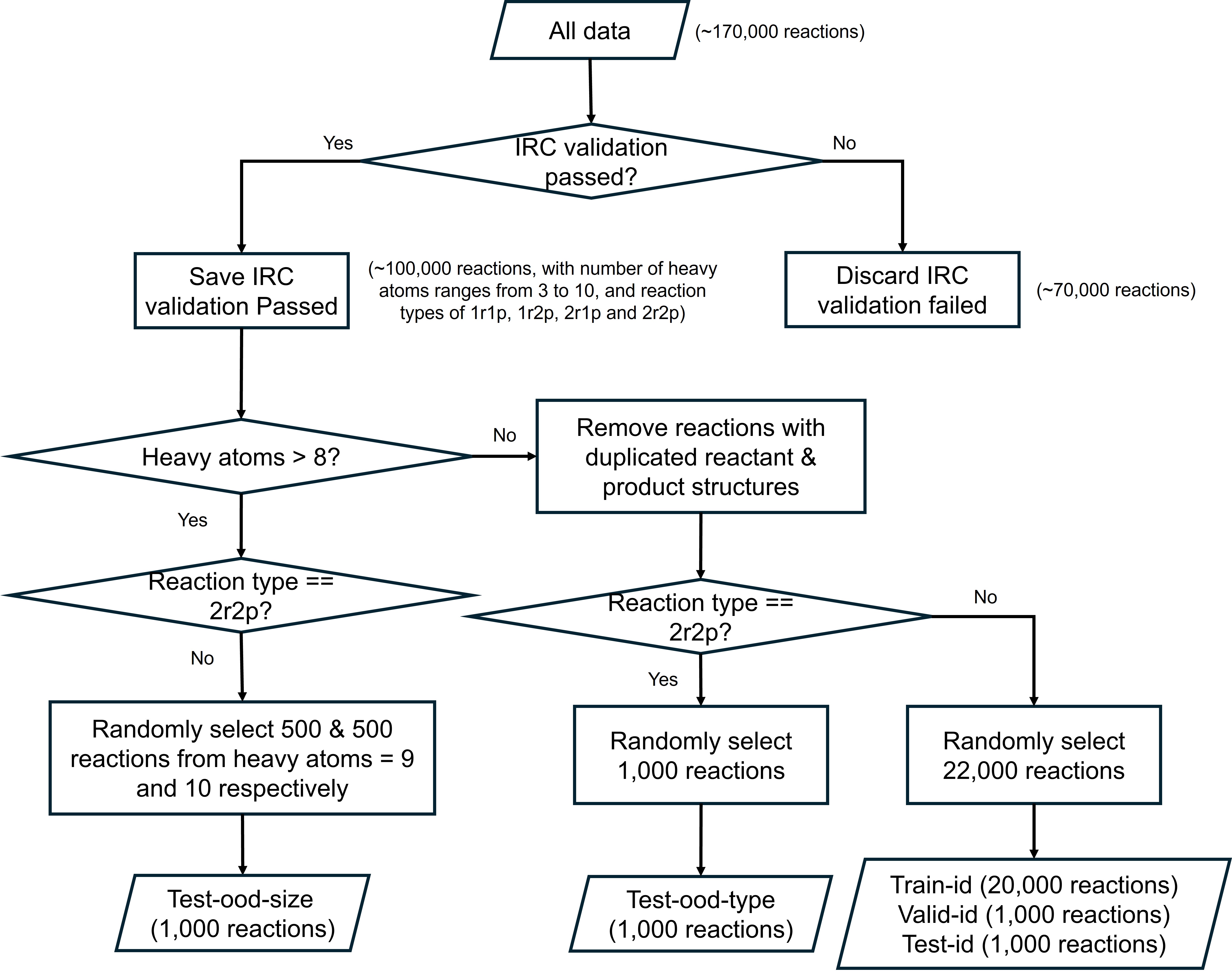}
    \caption{\textbf{The data splitting process of RGD1 dataset}.}
    \label{supp_fig:rgd1_split}
\end{figure}

\begin{figure}[ht]
    \centering
    \includegraphics[width=\textwidth]{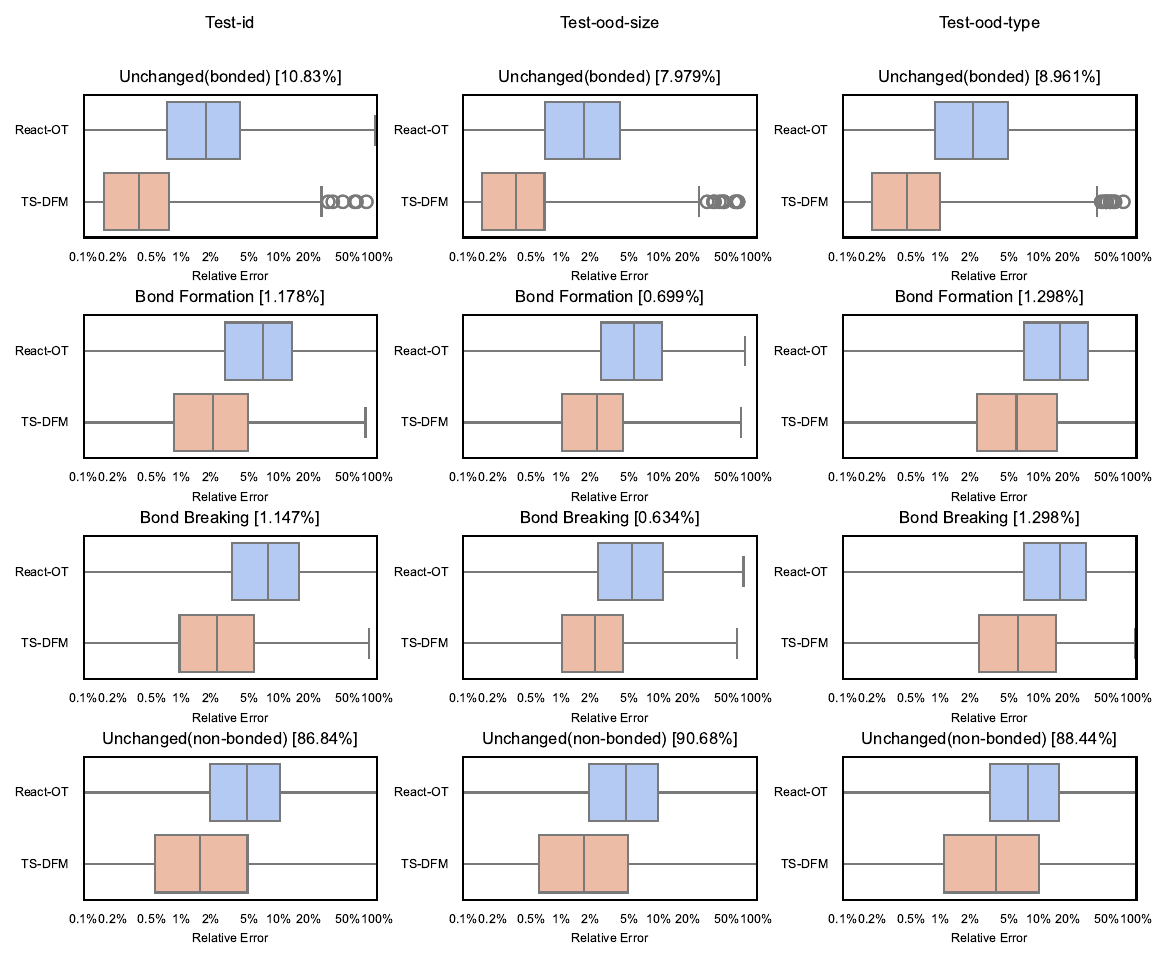}
    \caption{\textbf{Boxplots for the absolute percentage errors in pairwise distances of predicted TSs on RGD1 dataset, categorized by the type of bond change during the chemical reaction}.
    Percentages in brackets indicate the proportion of each bond change type within the entire test set.
    For all the bond change types, TS-DFM achieve superior performance than React-OT.
    }
    \label{supp_fig:rgd1_bond_analysis1}
\end{figure}

\begin{figure}[ht]
    \centering
    \includegraphics[width=7cm]{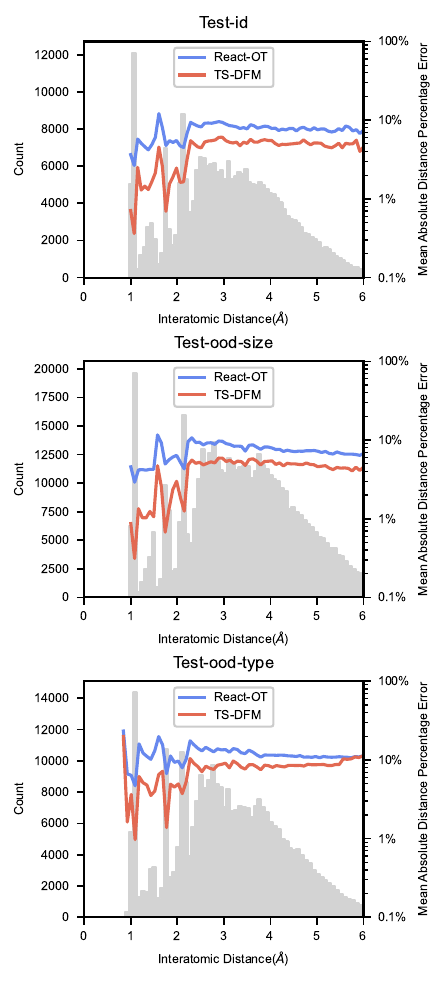}
    \caption{\textbf{Average absolute percentage errors and numbers of atom pairs w.r.t. pairwise distances on RGD1 dataset}.
    The gray histogram shows the distribution of interatomic distances.
    The colored line represents the mean absolute percentage error computed over different distance intervals.
    On three subsets, TS-DFM exhibits lower average error across almost all the interatomic distances.
    }
    \label{supp_fig:rgd1_bond_analysis2}
\end{figure}

\begin{figure}[ht]
    \centering
    \includegraphics[width=\textwidth]{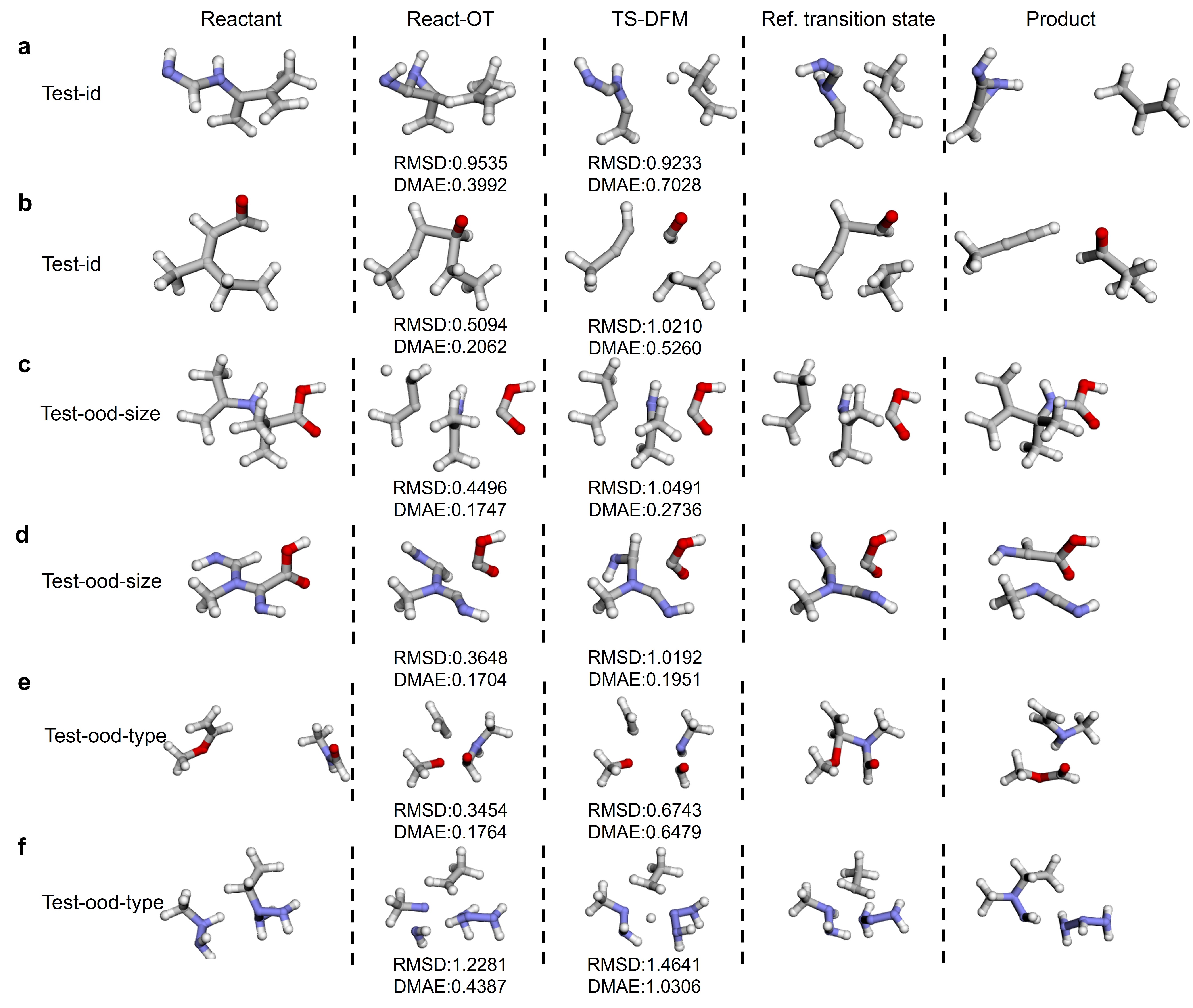}
    \caption{\textbf{More examples of React-OT outperforming TS-DFM on RGD1 dataset}.
    The H, C, N and O atoms are colored as white, gray, blue and red respectively.
    \textbf{a}, TS-DFM predicts the correct reaction mechanism; however, large RMSD and DMAE arise from the distance between the dissociated substructures.
    \textbf{b}, The large error of TS-DFM is caused by the misorientation of dissociated substructures.
    \textbf{c}, The large RMSD of TS-DFM originates from a reason similar to that in \textbf{b}.
    \textbf{d}, Incorrect torsion angles in the prediction of TS-DFM result in a large RMSD.
    \textbf{e}, Both React-OT and TS-DFM fail to predict the correct bondings in the TS.
    \textbf{f}, Similar to the case in \textbf{c}, the bond breakage is predicted successfully, but incorrect orientations of the dissociated substructures lead to large RMSD and DMAE values. 
    }
    \label{supp_fig:rgd1_reactot}
\end{figure}

\begin{figure}[ht]
    \centering
    \includegraphics[width=\textwidth]{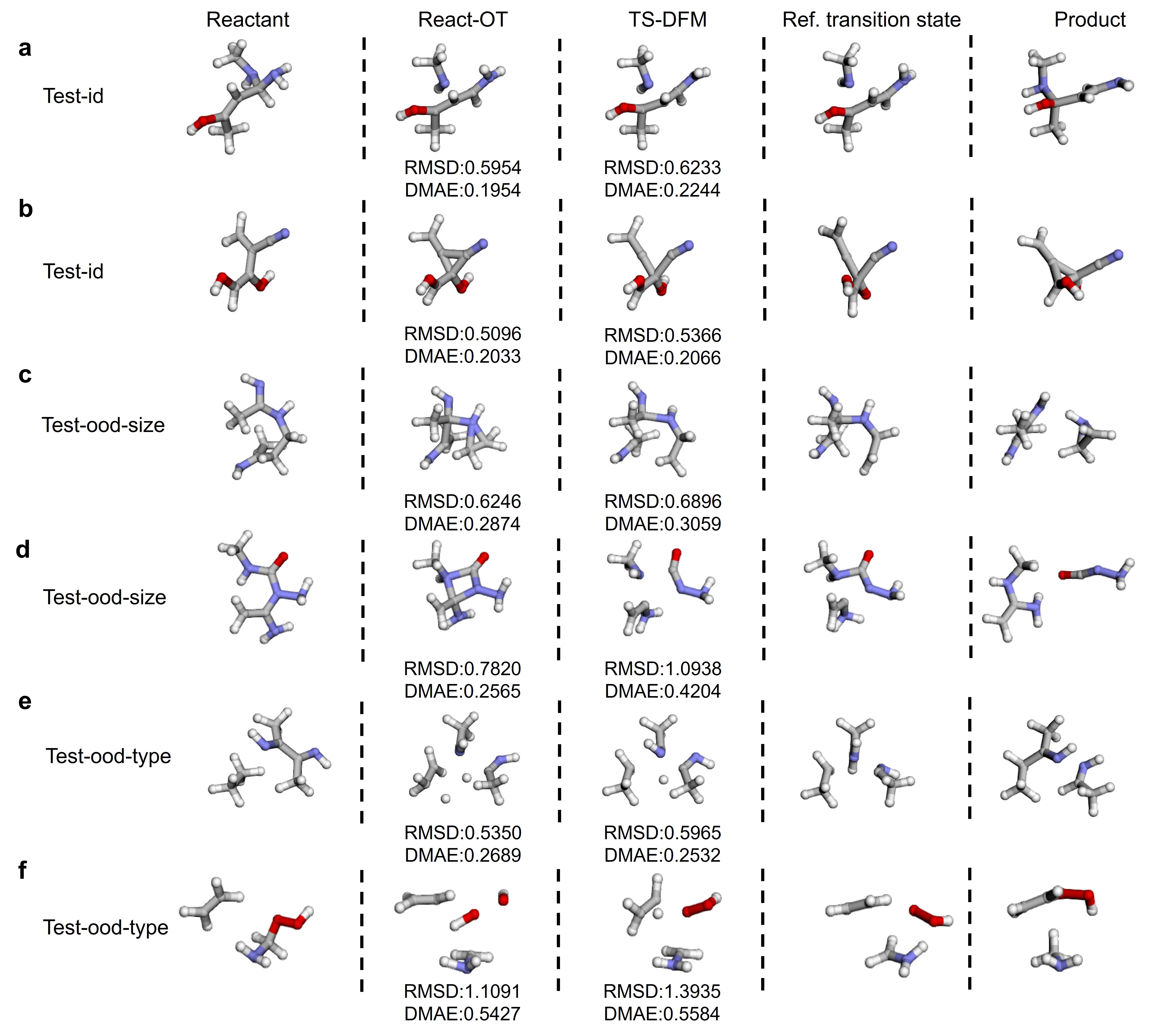}
    \caption{\textbf{More examples where React-OT has comparable performance with TS-DFM on RGD1 dataset}.
    The H, C, N and O atoms are colored as white, gray, blue and red respectively.
    \textbf{a}, React-OT and TS-DFM predict similar TSs.
    \textbf{b} and \textbf{c}, TS-DFM predicts the correct bondings in the predicted TS, while React-OT does not.
    \textbf{d}, \textbf{e} and \textbf{f}, TS-DFM produces TSs with bond cleavages and formations that are chemically relevant to the reaction process, although they contain incorrect patterns in comparison with true TSs.
    In contrast, the predictions of React-OT exhibit unexpected bond cleavages and formations.
    }
    \label{supp_fig:rgd1_comparable}
\end{figure}

\begin{figure}[ht]
    \centering
    \includegraphics[width=\textwidth]{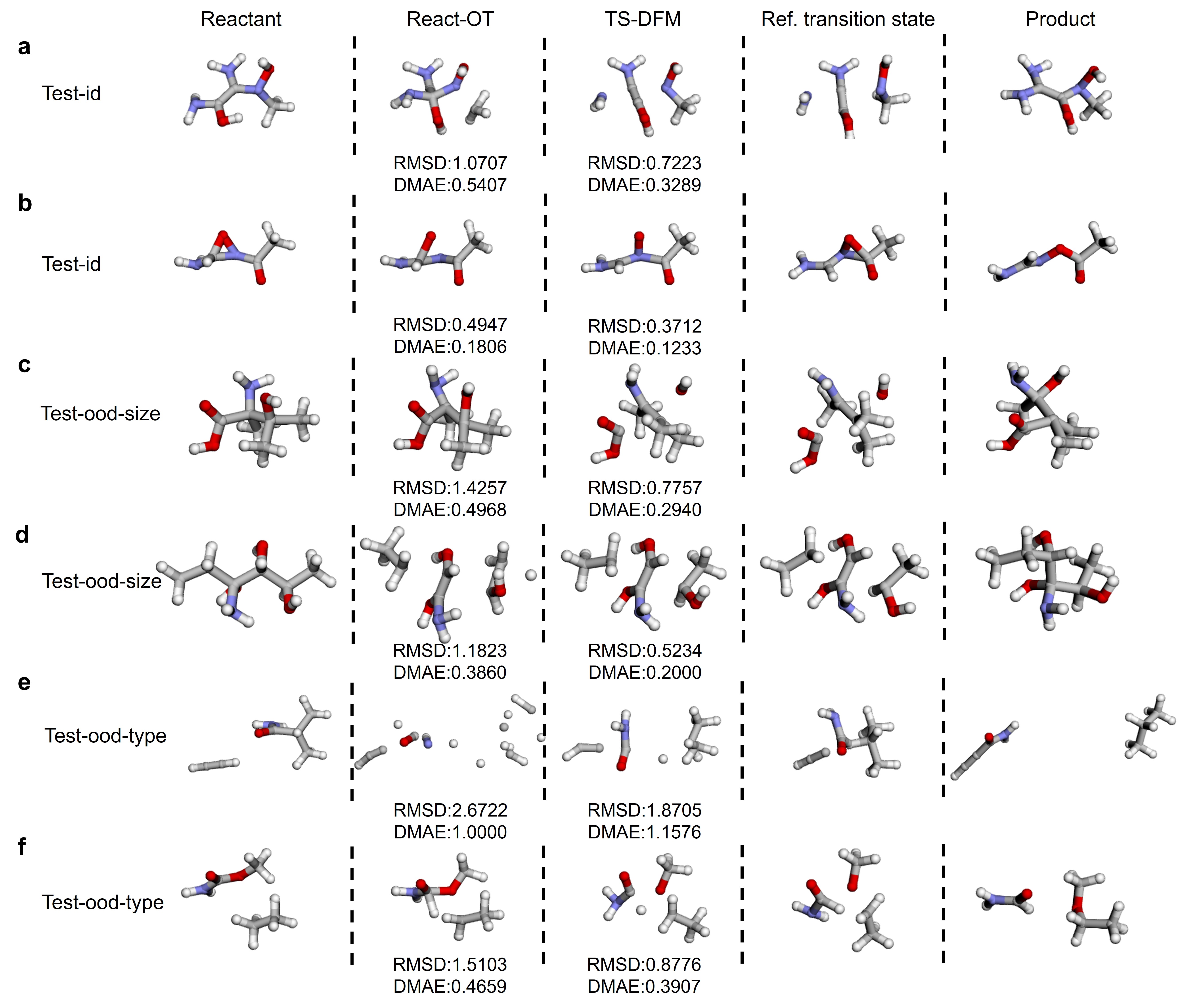}
    \caption{\textbf{More examples of TS-DFM outperforms React-OT on RGD1 dataset}.
    The H, C, N and O atoms are colored as white, gray, blue and red respectively.
    \textbf{a}, TS-DFM predicts the correct reaction mechanisms, whereas React-OT fails.
    \textbf{b}, Although both approaches fail to predict the exact TS structure, the prediction from TS-DFM is closer to the true TS geometry.
    \textbf{c}, The prediction of React-OT remains almost identical to the reactant structure, while TS-DFM successfully predicts the correct reaction mechanism.
    \textbf{d}, Although React-OT predicts the key bond cleavages as TS-DFM, the substructure misorientation and unexpected hydrogen dissociation result in a large RMSD.
    \textbf{e}, TS-DFM predicts a TS closer to the product, while the prediction of React-OT even scatters into separate atoms.
    \textbf{f}, React-OT only correctly predicts the transfer of hydrogen atom, while TS-DFM predicts the entire reaction mechanism correctly.
    }
    \label{supp_fig:rgd1_tsdfm}
\end{figure}

\clearpage

\section{Pseudocode for TS-DFM Training and Inference}
\begin{algorithm}[ht]
	\caption{Pseudocode for TS-DFM Training}
	\begin{algorithmic}
		\State \textbf{Input:}  training dataset  $\left[(Z^0,\vec{R}_{\text{R}}^0, \vec{R}_{\text{P}}^0,\vec{R}_{\text{TS}}^0),\dots, (Z^N,\vec{R}_{\text{R}}^N, \vec{R}_{\text{P}}^N,\vec{R}_{\text{TS}}^N)\right]$; initialized flow network $v_{\boldsymbol{\theta}}$; learning rate $\alpha$; noise scale $\sigma$.
		\State \textbf{Output:} trained flow network $v_{\boldsymbol{\theta}}$

        \While{training}

            \State sample $t\in \mathcal{U}[0,1]$ 
            \State sample $(Z,\vec{R}_{\text{R}}, \vec{R}_{\text{P}},\vec{R}_{\text{TS}})\in \left[(Z^0,\vec{R}_{\text{R}}^0, \vec{R}_{\text{P}}^0,\vec{R}_{\text{TS}}^0),\dots, (Z^N,\vec{R}_{\text{R}}^N, \vec{R}_{\text{P}}^N,\vec{R}_{\text{TS}}^N)\right]$

            \State $D_\text{R}=\sqrt{\text{diag}(\vec{R}_{\text{R}}\vec{R}_{\text{R}}^T)\boldsymbol{1}^T+\boldsymbol{1}\text{diag}(\vec{R}_{\text{R}}\vec{R}_{\text{R}}^T)^T-2\vec{R}_{\text{R}}\vec{R}_{\text{R}}^T}$
            \State $D_\text{P}=\sqrt{\text{diag}(\vec{R}_{\text{P}}\vec{R}_{\text{P}}^T)\boldsymbol{1}^T+\boldsymbol{1}\text{diag}(\vec{R}_{\text{P}}\vec{R}_{\text{P}}^T)^T-2\vec{R}_{\text{P}}\vec{R}_{\text{P}}^T}$
            \State $D_{\text{TS},1}=\sqrt{\text{diag}(\vec{R}_{\text{TS}}\vec{R}_{\text{TS}}^T)\boldsymbol{1}^T+\boldsymbol{1}\text{diag}(\vec{R}_{\text{TS}}\vec{R}_{\text{TS}}^T)^T-2\vec{R}_{\text{TS}}\vec{R}_{\text{TS}}^T}$
            \State $D_{\text{TS},0}=(D_\text{R}+D_\text{P})/2$
            \State $u_t\leftarrow D_{\text{TS},1}-D_{\text{TS},0}$
            \State sample $D_{\text{TS},t}\sim \mathcal{N}(D_{\text{TS},t}|t\cdot D_{\text{TS},1}+(1-t)\cdot D_{\text{TS},0}, \sigma)$
            \State $\mathcal{L}_{\text{TS-DFM}}=\|u_t-v_{\boldsymbol{\theta}}(D_{\text{TS},t},t|Z,D_{\text{R}},D_{\text{P}})\|^2$
            \State $\boldsymbol{\theta}\leftarrow \boldsymbol{\theta}-\alpha \nabla_{\boldsymbol{\theta}}\mathcal{L}_{\text{TS-DFM}}$
            
        \EndWhile
        
		\State \Return $v_{\boldsymbol{\theta}}$
	\end{algorithmic}
    \label{supp_alg:training}
\end{algorithm}

\begin{algorithm}[ht]
	\caption{Pseudocode for TS-DFM Inference}
	\begin{algorithmic}
		\State \textbf{Require:} trained flow network $v_{\boldsymbol{\theta}}$; step size $\Delta t$; atom type vector $Z$; pairwise distance matrices of reactant and product $D_{\text{R}},D_{\text{P}}$.
        \State $\hat{D}_{\text{TS}}=(D_\text{R}+D_\text{P})/2$
        \State $t=0$
        \While{$t\le 1$}
            \State $\hat{v}=v_{\boldsymbol{\theta}}\left(\hat{D}_{\text{TS}},t|Z,D_{\text{R}},D_{\text{P}}\right)$
            \State $\hat{D}_{\text{TS}}\leftarrow \hat{D}_{\text{TS}}+\hat{v}\cdot\Delta t$
            \State $t\leftarrow t+\Delta t$
            
        \EndWhile
        \State $\hat{R}_{\text{TS}}=\texttt{MultiDimensional\_Scaling}(\hat{D}_{\text{TS}}, 3)$
        \State $\hat{R}_{\text{TS}}=\arg\min_{\hat{R}_{\text{TS}}} \left[\left({\text{diag}(\hat{R}_{\text{TS}}\hat{R}_{\text{TS}}^T)\boldsymbol{1}^T+\boldsymbol{1}\text{diag}(\hat{R}_{\text{TS}}\hat{R}_{\text{TS}}^T)^T-2\hat{R}_{\text{TS}}\hat{R}_{\text{TS}}^T}\right)-\hat{D}_{\text{TS}}^2\right]/\hat{D}_{\text{TS}}^2$
		\State \Return $\hat{R}_{\text{TS}}$,
	\end{algorithmic}
    \label{supp_alg:inference}
\end{algorithm}

\clearpage
\bibliography{supplementary.bib}